\documentclass[twoside,leqno,twocolumn]{article}

\usepackage[letterpaper]{geometry}

\usepackage{ltexpprt}
\usepackage{hyperref}

\usepackage{cite}
\usepackage{amsmath,amssymb,amsfonts}
\usepackage[algo2e, ruled, linesnumbered]{algorithm2e}
\usepackage{graphicx}
\usepackage{textcomp}
\usepackage{xcolor}
\def\BibTeX{{\rm B\kern-.05em{\sc i\kern-.025em b}\kern-.08em
    T\kern-.1667em\lower.7ex\hbox{E}\kern-.125emX}}

\usepackage{float}
\usepackage{subcaption}
\usepackage[normalem]{ulem}
\usepackage{booktabs}

\usepackage{url}
\usepackage{multirow}
\usepackage{diagbox}
\usepackage{textcomp}

\usepackage{fix-cm}
\usepackage{xfrac}
\usepackage{makecell}
\usepackage{booktabs}

\newcommand{\Methoddip}{Dip'n'Sub\xspace}
\newcommand{\diptest}{Dip-test\xspace}
\newcommand{\dipval}{Dip-value\xspace}
\newcommand{\dippvalue}{Dip-p-value\xspace}

\begin{document}

\title{\Large Extension of the Dip-test Repertoire - Efficient and Differentiable p-value Calculation for Clustering}

\author{Lena G. M. Bauer\thanks{Authors contributed equally.}~$^,$\footnote{
Faculty of Computer Science, ds:UniVie, University of Vienna, Vienna, Austria. \{lena.bauer, christian.boehm, claudia.plant\}@univie.ac.at}~$^,$\footnote{UniVie Doctoral School Computer Science.}
\and Collin Leiber$^{\ast,}$\thanks{LMU Munich \& MCML, Munich, Germany. leiber@dbs.ifi.lmu.de}     
\and Christian Böhm$^\dagger$
\and Claudia Plant$^\dagger$
}

\date{}

\maketitle

\fancyfoot[R]{\scriptsize{Copyright \textcopyright\ 2023 by SIAM\\
Unauthorized reproduction of this article is prohibited}}

\begin{abstract} \small\baselineskip=9pt 
Over the last decade, the Dip-test of unimodality has gained increasing interest in the data mining community as it is a parameter-free statistical test that reliably rates the modality in one-dimensional samples. It returns a so called Dip-value and a corresponding probability for the sample's unimodality (\dippvalue). These
two values share a sigmoidal relationship. However, the specific transformation is dependent on the sample size. Many Dip-based clustering algorithms use bootstrapped look-up tables translating Dip- to Dip-p-values for a certain limited amount of sample sizes. 
We propose a specifically designed sigmoid function as a substitute for these state-of-the-art look-up tables. This accelerates computation and provides an approximation of the Dip- to Dip-p-value transformation for every single sample size.
Further, it is differentiable and can therefore easily be integrated in learning schemes using gradient descent. We showcase this by exploiting our function in a novel subspace clustering algorithm called Dip'n'Sub. We highlight in extensive experiments the various benefits of our proposal.
\end{abstract}

\section{Introduction}

\begin{figure}[t]
\centering
\begin{subfigure}[t]{0.94\linewidth}
\includegraphics[width=\linewidth]{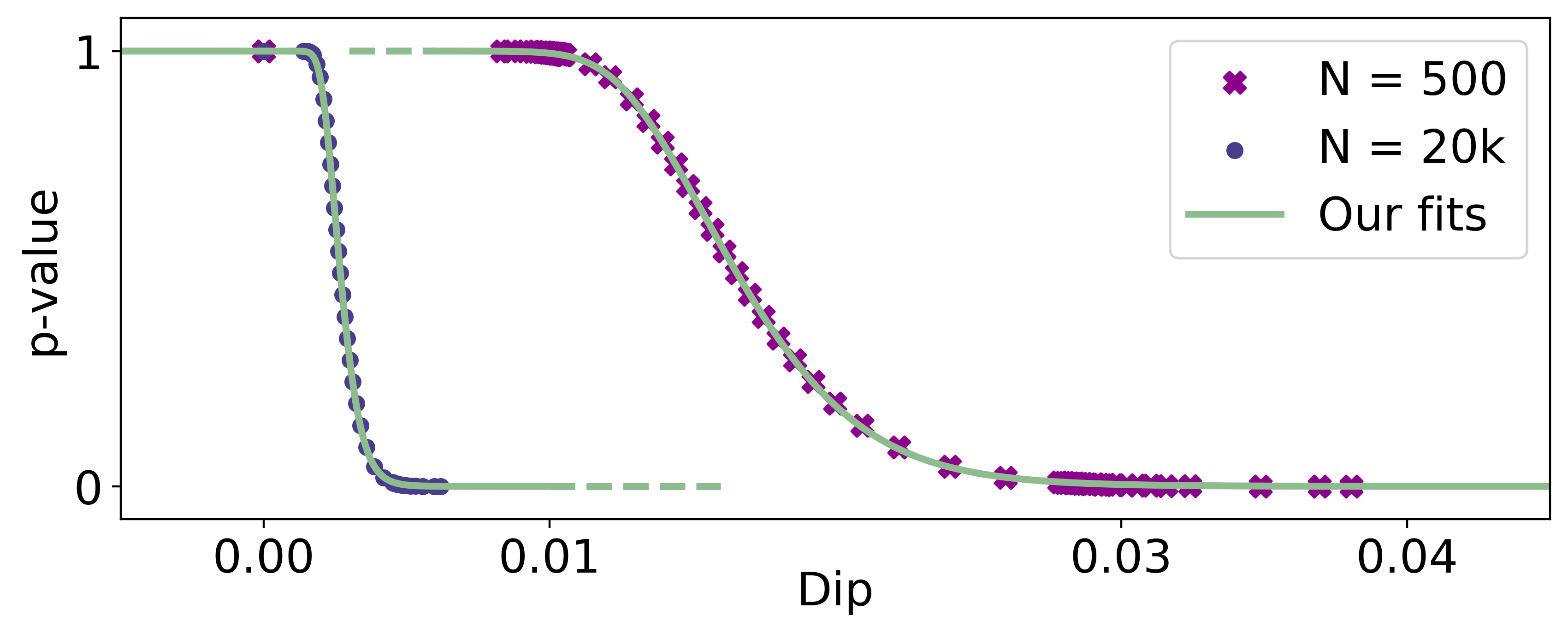}
\captionsetup{format=hang}
\subcaption{Bootstrapped ($Dip$,$p$)-pairs and our fitted functions.}
\label{fig:boot_intro}
\end{subfigure}\\
\begin{subfigure}[t]{0.465\linewidth}
\includegraphics[width=\linewidth]{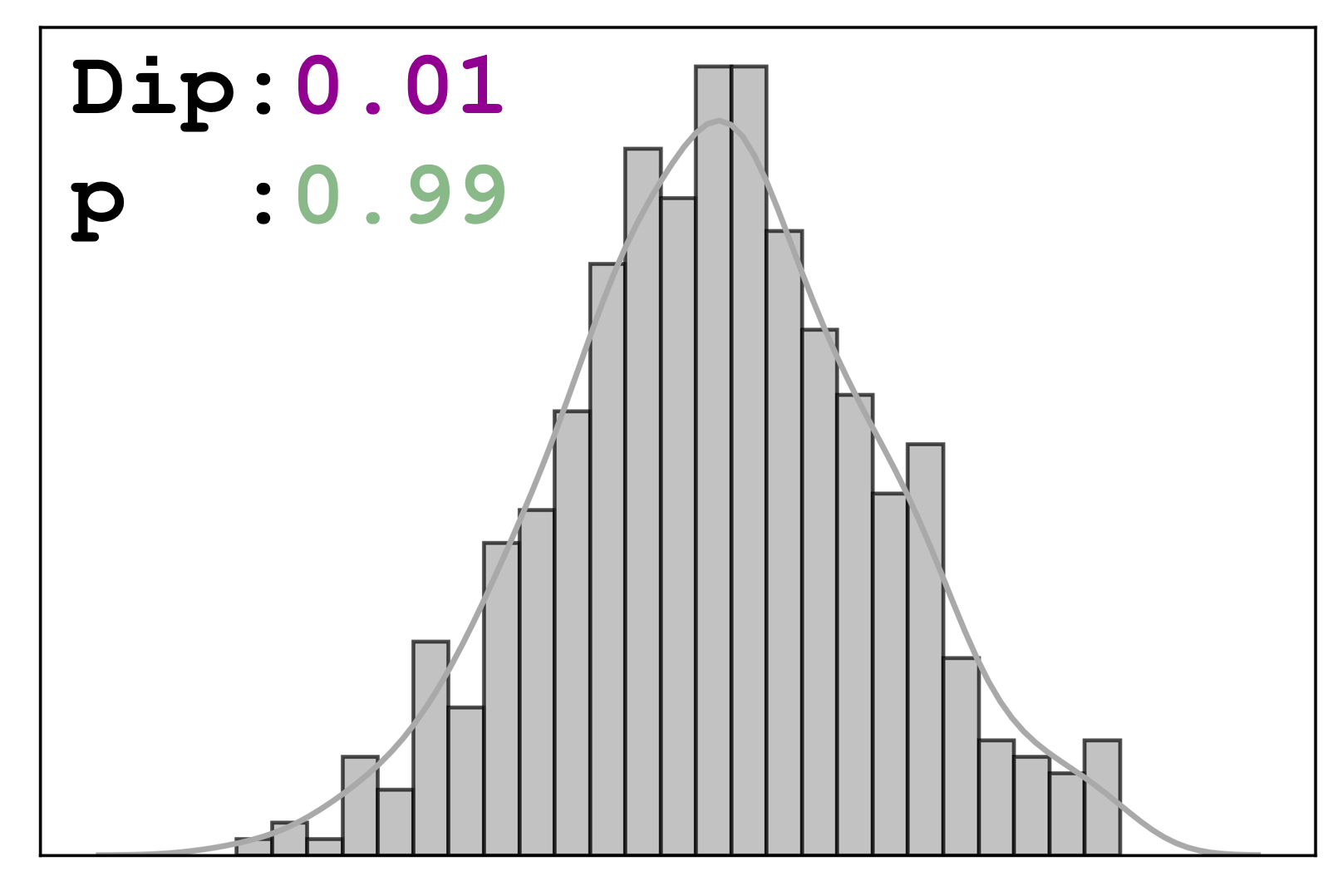}
\captionsetup{format=hang}
\subcaption{$X\sim \mathcal{N}(0,1)$, $N = 500$.}
\label{fig:parameters_gaussian_500}
\end{subfigure}
\begin{subfigure}[t]{0.465\linewidth}
\includegraphics[width=\linewidth]{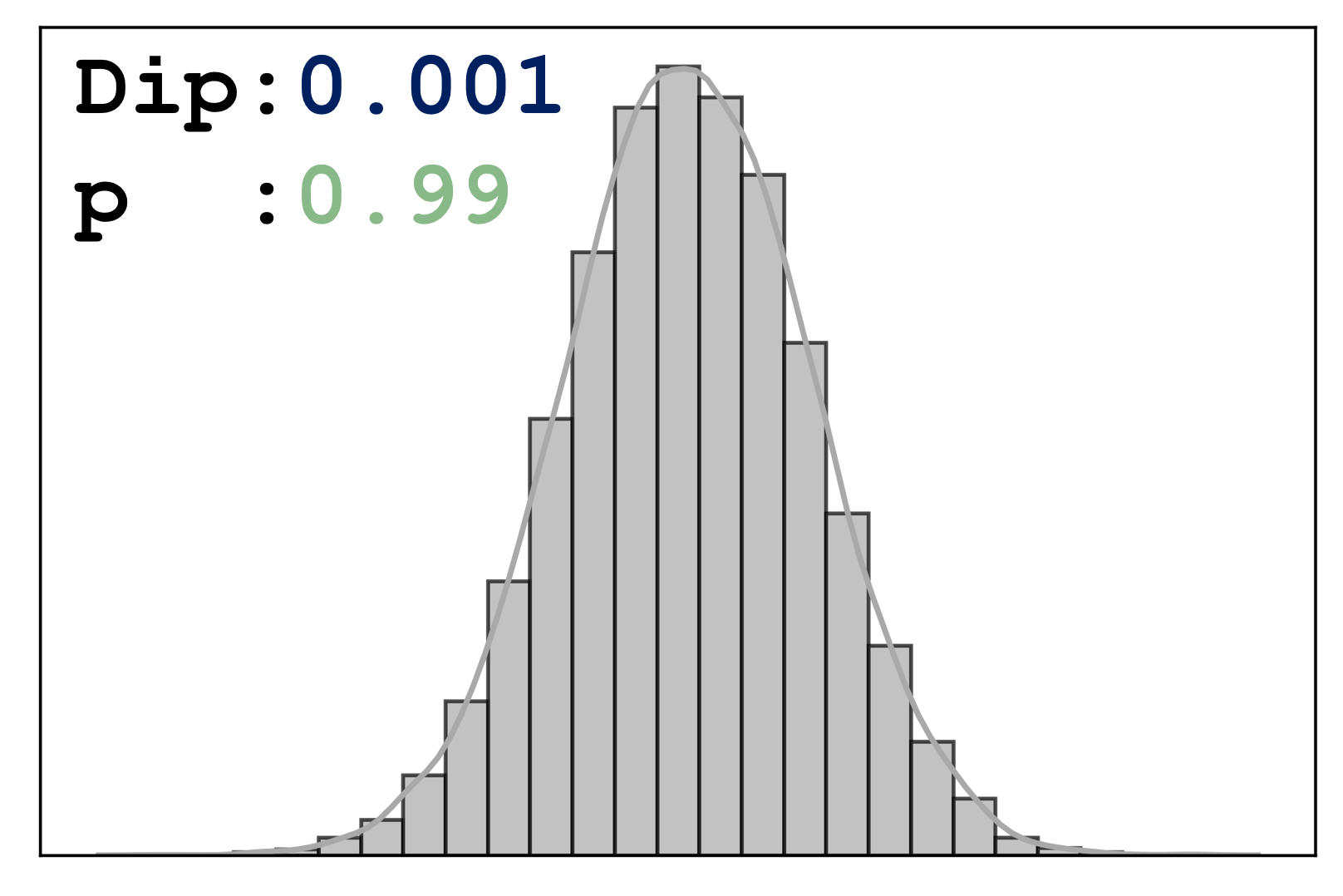}
\captionsetup{format=hang}
\subcaption{$X \sim \mathcal{N}(0,1)$, $N = 20k$.}
\label{fig:gaussian_20k}
\end{subfigure} \hfill
\caption{\textbf{(a)}
Bootstrapped $(Dip,p)$-pairs for sample sizes $N = 500$ (purple) and $N = 20k$ (blue) and our fitted functions (green). The transformation function from Dip- to \dippvalue strongly depends on the sample size. \textbf{(b)} and \textbf{(c)} Histograms of samples from a $\mathcal{N}(0,1)$ normal distribution with sample size $N = 500$ and $N = 20k$. When applying the \diptest on the two samples, their Dip-values differ with a factor of $10$. 
The respective Dip-p-values are, however, $0.99$ in both cases.
}
\label{fig:intro}
\end{figure}

One of the major goals in data mining is to automatically find meaningful patterns and structures in data. This is ideally done fast and with as few hyperparameters as possible. 
Here, the definition of statistical modes plays a decisive role in many approaches. Clustering methods like Meanshift \cite{meanshift} or Quickshift \cite{quickshift} try to find modes with the help of a predefined influence area and assign objects to these modes. Other procedures focus on evaluating whether a data set has a single or multiple modes. Examples are Hartigan's \diptest \cite{hartigan}, the Silverman test \cite{silvermantest}, the Folding Test \cite{gathereddata} or the recently presented UU-test \cite{uutest}.
In clustering the assumption for these methods is that multiple modes indicate multiple clusters, while unimodality is a sign for a single cluster. The tests are basically parameter-free, which makes them particularly interesting for the data mining community. The most commonly used unimodality test in the clustering domain is probably the \diptest.
The input for the \diptest is a one-dimensional sample and the test outputs a \dipval $Dip \in (0,0.25]$ that can be converted to a $p$-value, which we will term \dippvalue throughout this work. The latter represents a probability for the sample to be unimodal and it is high for Dip-values close to zero and low for Dip-values close to its maximum of $0.25$. Its benefits have already been exploited by several data mining techniques.

DipMeans \cite{dipmeans}, projected DipMeans \cite{p_dipmeans}, SkinnyDip \cite{skinnydip}, M-Dip \cite{mdip}, NrDipMeans \cite{nrdipmeans} and DipDECK \cite{dipdeck}, for example, are all algorithms that use the \diptest to estimate the number of clusters. 
Other algorithms like DipTransformation \cite{diptrans}, DipExt \cite{dipext} or the DipEncoder \cite{dipencoder} utilise the \diptest to create cluster-friendly lower-dimensional spaces.
When executing these algorithms, a precise and efficient determination of the Dip- and \dippvalue is of high relevance. 
In general, the \dipval translates sigmoidal to the \dippvalue. 
Figure \ref{fig:boot_intro} clearly shows the pattern. Importantly, we can observe that the specific curvatures and positions of the sigmoidal functions are heavily dependent on the number of samples $N$. Figures \ref{fig:parameters_gaussian_500} and \ref{fig:gaussian_20k} show that the Dip-p-values $p$ of a normal distribution $\mathcal{N}(0,1)$ with $N=500$ and $N=20k$ are both $0.99$. This is in accordance to the expectation of the normal distribution to be unimodal irrespective of the sample size. The Dip-values $Dip$, however, differ by a factor of $10$. Thus, considering the \dippvalue is a far more robust choice when optimising for uni- or multimodality. 

To the best of our knowledge, however, there exists no elegant method for the translation from Dip- to \dippvalue. The state-of-the-art procedure is to utilise a look-up table with a limited amount of pairs of Dip- and Dip-p-values and use $\sqrt{N}$-interpolation to provide the missing pairs in between table values. While this approach has been used in multiple procedures (e.g., \cite{dipdeck, skinnydip, dipmeans, p_dipmeans, nrdipmeans}), it shows drawbacks in several aspects. First, it is not differentiable and therefore harder to integrate into strategies such as stochastic gradient descent (SGD) and second, it is limited to the maximum bootstrapped sample size in the table. 
To resolve these shortcomings we propose a differentiable transformation function which - provided with a \dipval and the number of samples - automatically calculates the corresponding \dippvalue (see green lines in Fig. \ref{fig:boot_intro}). We showcase the practical value of our proposal for data mining research by developing a subspace clustering algorithm, that exploits the differentiability of our transformation function to identify a common subspace for all clusters in the data set. This helps to analyse relationships between clusters \cite{fossclu, subkmeans} and distinguishes us from `classical' subspace clustering algorithms (e.g., ORCLUS \cite{orclus} or 4C \cite{4c}) which find a separate subspace for each cluster. Our idea uses the gradient of the \dippvalue to identify projection axes where all clusters show a high degree of multimodality by performing SGD. This approach makes it particularly apt to rely on the \dippvalue instead of the \dipval as different cluster sizes bias the latter. Further, we present TailoredDip, an extended version of the one-dimensional clustering procedure UniDip \cite{skinnydip}, to cluster the data on those axes.

Our contributions can be summarised as follows:
\begin{itemize}
    \item We provide a fully automatic translation from Dip- to \dippvalue via an analytical function (Sec. \ref{sec:methodsfit})
    \item The translation is available for any data size $N$ and provides reliable Dip-p-values irrespective of the underlying distribution (Sec. \ref{sec:reliability_consistency})
    \item Analyses show that our novel calculation is faster than previously used methods, in particular bootstrapping (Sec.\ref{sec:runtime})
    \item We showcase how the differentiability of our function is useful for practical data mining applications by introducing our subspace clustering algorithm \Methoddip (Sec. \ref{sec:dipnsub} and \ref{sec:subspace_res})
\end{itemize}

\section{Related Work}
\subsection{The Dip-test}
\label{sec:the-dip-test}
The \diptest is a statistical test for modality in a one-dimensional sample developed by Hartigan and Hartigan \cite{hartigan}. The test returns a so-called \dipval $Dip$ that specifies the distance between the Empirical Cumulative Distribution Function (ECDF) of the sample to an unimodal piece-wise linear function, i.e., a function that is convex up to the beginning of the steepest slope and concave thereafter. In this context, the area of steepest slope is often called \textit{modal interval}. By definition $Dip \in (0,0.25]$, where a $Dip$ close to $0$ indicates a unimodal distribution and a $Dip \gg 0$ indicates a multimodal distribution. The exact value not only depends on the specific distribution but also on the sample size (see Fig. \ref{fig:parameters_gaussian_500} and \ref{fig:gaussian_20k}). For this reason, Dip-values are often not used directly, but the associated Dip-p-values.
Here, the null hypothesis $\mathcal{H}_0$ states that the sample set is unimodal and the alternative hypothesis $\mathcal{H}_1$ is that there are at least two modes present in the data. 
The test does not make any assumptions about the data generating distribution. For any distribution with a single mode, whether it is Gaussian, Laplacian or t-distributed the test will not reject $\mathcal{H}_0$. The \diptest can be calculated efficiently in $O(N)$ \cite{hartigan} on a sorted input of size $N$. Furthermore, \cite{krause2005multimodal} showed that the \dipval is differentiable with respect to a projection axis. This enables the use of SGD. Due to these several benefits, the \diptest has successfully been integrated in several data mining applications over the last decade. 

\subsection{Bootstrapping}
The \diptest can convert a \dipval to a $p$-value representing an evidence measure for the credibility of the null hypothesis. The original work by Hartigan and Hartigan \cite{hartigan} provides a table with bootstrapped Dip-values and corresponding Dip-p-values for samples of different sizes $N$. 
The table lists the relationships for $13$ different sample sizes ranging between $4$ and $200$ with $9$ $(Dip,p)$-pairs each. These were calculated by drawing $N$ random samples from a uniform distribution several times. The \diptest was then performed on each of these sample sets. The percentage of sets with a \dipval smaller than the input \dipval thus gives the respective \dippvalue. The idea is that the uniform distribution represents a borderline case between unimodal and multimodal distributions \cite{hartigan}.
In previous publications authors used bootstrapping to enlarge Hartigan and Hartigan's table to a total of $21$ sample sizes up to $72,000$ data points with $26$ ($Dip$, $p$)-pairs for each sample size. They use interpolation of $\sqrt{N}$ to get intermediate Dip-p-values. To the best of our knowledge this approach is the state-of-the-art when transforming Dip- to \dippvalue. 

\subsection{\diptest Related Data Mining}
Most clustering methods utilising the \diptest are dedicated to the estimation of the number of clusters in a data set. Here, an essential component is the transformation of the data into a one-dimensional space where the \diptest can be applied. One of the first approaches is DipMeans \cite{dipmeans}, which uses the \diptest to determine if the distances between data points within a cluster are distributed unimodally. For projected DipMeans \cite{p_dipmeans}
the input for the \diptest are the data points itself, after they have been projected onto projection axes. Another transformation is applied by M-Dip \cite{mdip}. Here, the \diptest is executed on the path of closely located points between two clusters. NrDipMeans \cite{nrdipmeans} estimates the number of clusters in a non-redundant clustering setting by employing a Dip-based splitting strategy. 
The first clustering and $k$-estimation technique using the \diptest in a deep learning context is DipDECK \cite{dipdeck}, which uses the test to decide whether clusters should be merged or not.
SkinnyDip \cite{skinnydip} determines clusters in highly noisy data sets by running its subroutine UniDip on each feature of each cluster. UniDip recursively executes the \diptest to identify modal intervals (see Sec. \ref{sec:the-dip-test}) in the one-dimensional data set until all intervals themselves and the areas left and right (until the next interval) are considered unimodal. Finally, each modal interval is considered a cluster and all objects that do not fall within such an area are classified as noise.
The idea of SkinnyDip is later used to cluster streaming data with StrDip \cite{StrDip}.
All of the mentioned procedures need to decide whether samples are distributed unimodally or not and, therefore, use the same bootstrapped look-up table to convert Dip-values into probabilities. 

An example for a Dip-based pre-processing algorithm is DipTransformation \cite{diptrans}. 
It scales and transforms a data set, such that the resulting feature space is suitable for k-means. In DipExt \cite{dipext} this idea is extended by making use of the differentiability of the \dipval with respect to the projection axis. Structure-rich features are extracted from the data by searching for suitable projection axes with SGD. 
The DipEncoder \cite{dipencoder} uses the gradient of the \dipval to train an autoencoder in such a way that each combination of clusters projected onto their specific projection axis is highly multimodal in the embedding.

\subsection{Common Subspace Clustering}
Traditional subspace clustering algorithms like 4C \cite{4c} or ORCLUS \cite{orclus} define an individual subspace for each cluster. In this setting, however, the inter-cluster relationships are difficult to analyse \cite{fossclu}. We would therefore like to identify a common subspace for all clusters.
A simple possibility to find such subspaces are dimensionality reduction techniques such as Principal Component Analysis (PCA) \cite{pca}, Independent Component Analysis (ICA) \cite{ica}, Linear Discriminant Analysis (LDA) \cite{lda} or the already mentioned DipExt \cite{dipext} algorithm. 
Since in these cases possible cluster assignments do not influence the final subspace, special common subspace clustering algorithms like LDA-k-means \cite{ldakmeans}, FOSSCLU \cite{fossclu} and SubKmeans \cite{subkmeans} have been developed. 
LDA-k-means and SubKmeans utilise LDA or an eigenvalue decomposition, respectively, in combination with the k-means objective to determine the subspaces. FOSSCLU combines the EM algorithm with rigid transformations.

\section{Methods}
In this section, we design a special sigmoid function converting Dip- to Dip-p-values. This function's differentiability is then exploited by our subspace clustering algorithm \Methoddip.

\subsection{Table Extension}
\label{ssec:table_ext}
\begin{figure*}
\centering
\begin{subfigure}[t]{0.32\textwidth}
\includegraphics[width=\textwidth]{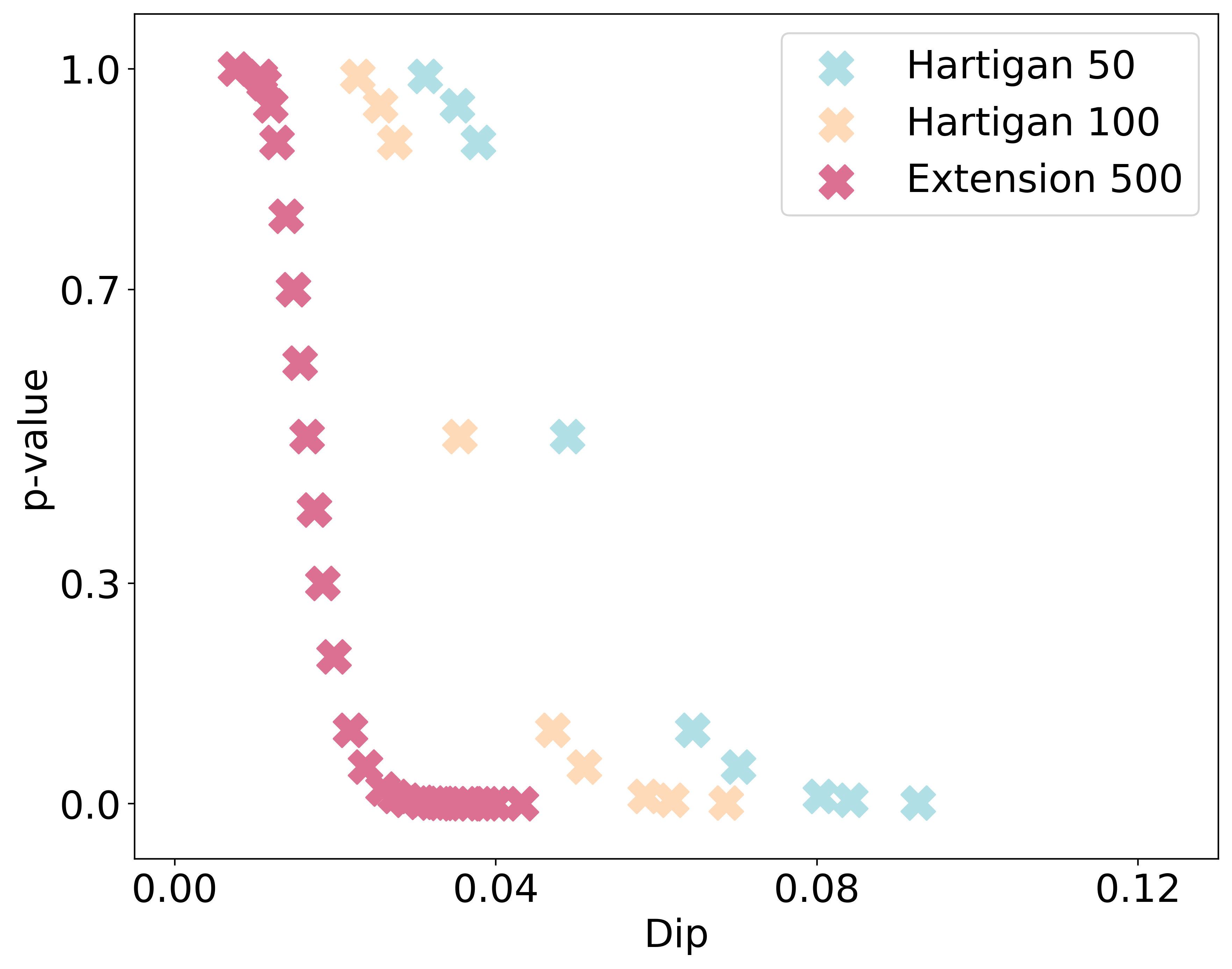}
\captionsetup{format=hang}
\subcaption{Examples of $(Dip,p)$-pairs from the original table \cite{hartigan} and an example of an extensions for $N=500$}
\label{fig:hartigan}
\end{subfigure}\hfill
\begin{subfigure}[t]{0.32\textwidth}
\includegraphics[width=\textwidth]{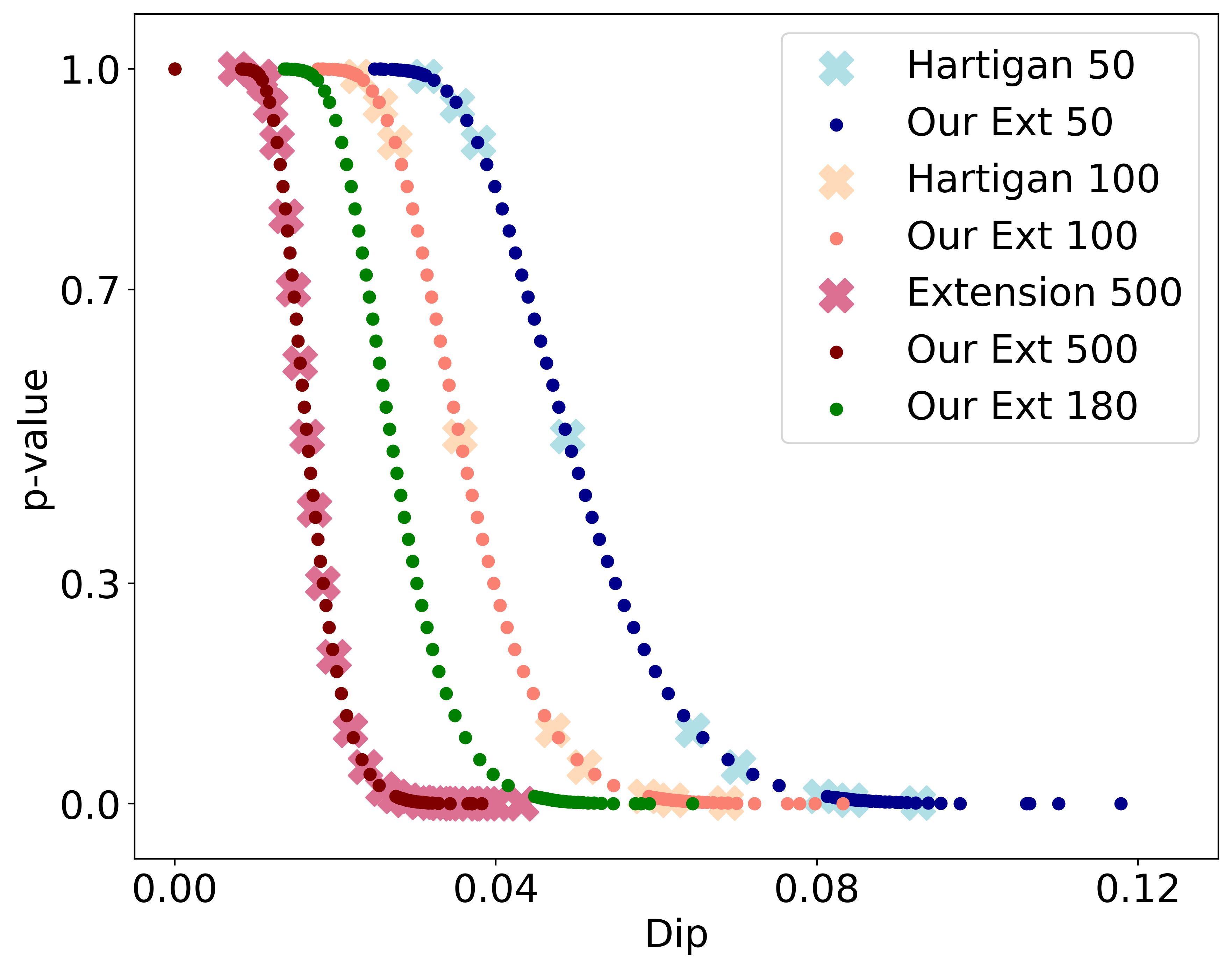}
\captionsetup{format=hang}
\subcaption{We bootstrap with larger granularity for $N$ as well as $(Dip,p)$-pairs.}
\label{fig:extension}
\end{subfigure}\hfill
\begin{subfigure}[t]{0.32\textwidth}
\includegraphics[width=\textwidth]{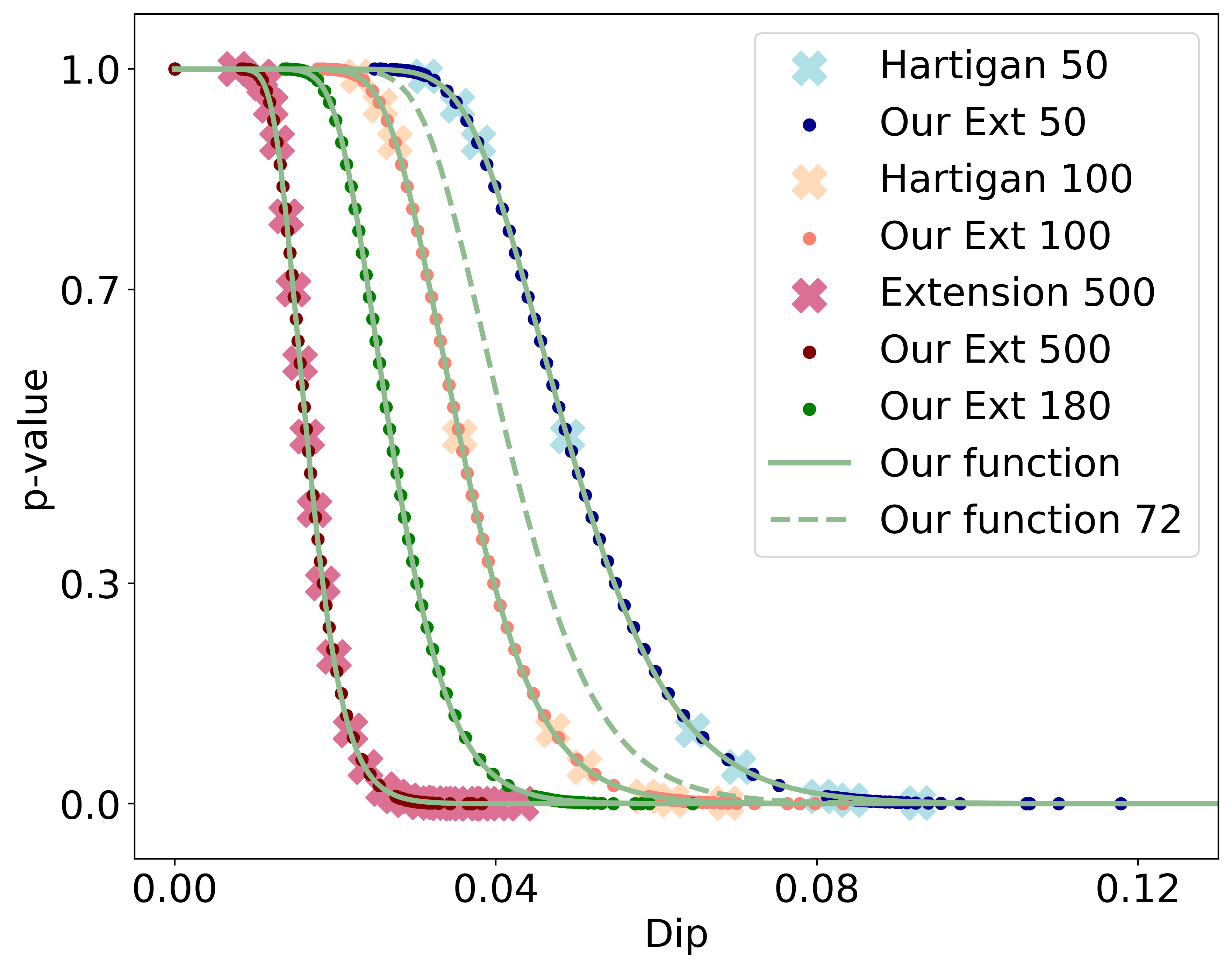}
\captionsetup{format=hang}
\subcaption{Our differentiable fitted function closes the remaining granularity gaps for all $N$ and all $(Dip,p)$-pairs.}
\label{fig:fits}
\end{subfigure}
\caption{\textbf{(a)} Hartigan and Hartigan's original bootstrapped table only provides pairs of Dip- and Dip-p-values for $13$ different sample sizes. This table has been extended to values of $N \geq 500$. \textbf{(b)} We enlarge the table for even more values of $N$ as well as a larger granularity regarding $(Dip,p)$-pairs (for better visualisation, we down-sampled our table to every third point). \textbf{(c)} We close the remaining gaps by providing our fitted function, such as for $N=72$, for which we do not have bootstrapped values.}
\label{fig:methods1}
\end{figure*}

To increase the granularity level of the look-up table previously used in literature, we also use bootstrapping as proposed by \cite{hartigan}. We sample from uniform distributions with $100$,$000$ repetitions to obtain a \dippvalue table containing $307$ $(Dip,p)$-pairs for $63$ sample sizes up to a sample size of $150$,$000$ data points. This look-up table provides a good basis, but it does not cover all values of sample sizes $N$.
Therefore, we fit a logistic function $p(\cdot)$ to provide a differentiable solution for this issue. Fig. \ref{fig:methods1} visually captures Sec. \ref{ssec:table_ext} and Sec. \ref{sec:methodsfit}.

\subsection{Function Fit}
\label{sec:methodsfit}
\begin{figure}[t]
\centering
\includegraphics[width=\linewidth]{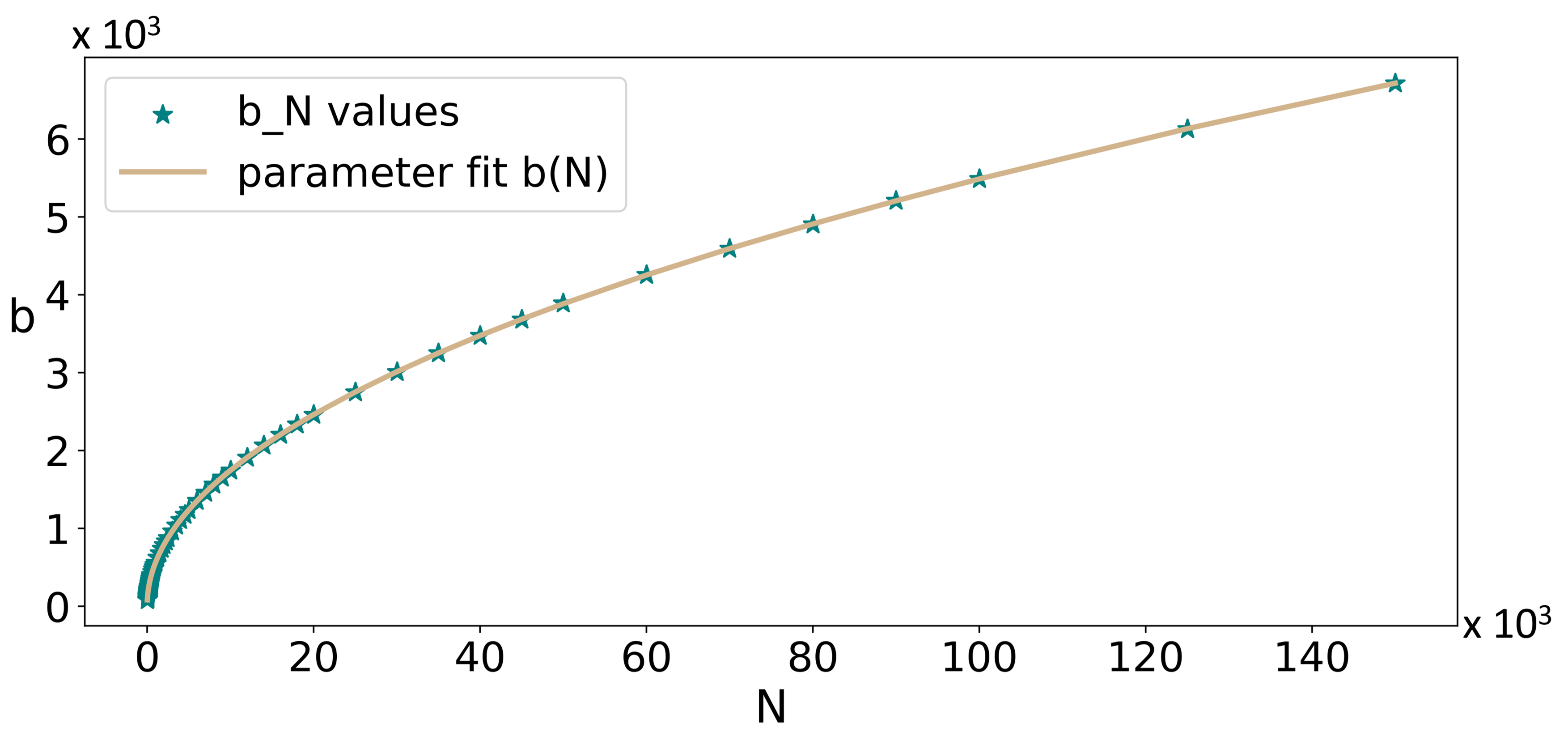}
\caption{Our fitted function for the parameter $b_N$.}
\label{fig:parameters_b}
\end{figure}

A good fitting function provides high flexibility to fit the different sigmoidal relationships between Dip- and \dippvalue for all sample sizes $N$ while using the smallest possible number of parameters. Due to the sigmoidal behaviour it is reasonable to approximate it with a generalised logistic function \cite{richardslogistic}.  
\begin{align*}
l(Dip) = d + \frac{a-d}{\left(c+h\cdot e^{-b\cdot Dip}\right)^{\sfrac{1}{g}}}
\end{align*}
The parameters $a$ and $d$ represent the upper and lower asymptote, respectively, and need to be fixed as $a=1$ and $d=0$ in our application as we model probability values. In addition, $c=1$ must hold for the function to actually be constrained by $1$. Further, often $h=g$ applies (see e.g. \cite{flexibleSigmoidalGrowth}). To obtain a negative slope, $b$ needs to be positive.
Another requirement is that $g>0$ holds. This essentially leaves the parameters $h=g$ and $b$, which mainly determine the curvatures. However, we also need to adjust the scale for the x-axis since the \dipval is limited within $(0,0.25]$. Therefore, we have to include a shift parameter in the exponential function. Finally, we want to account for highly different curvatures at the two asymptotes and therefore include a weighting, which is partly inspired by the work of \cite{Ricketts1999AFL}.
Our final fitting function reads as follows:
\begin{align*}
    p(x,\theta_N) &= 1 -
    \bigg[ w_N\cdot(1+h_N \cdot e^{-q_N\cdot x+s_N})^{\sfrac{1}{h_N}} \\
    &+  (1-w_N)\cdot(1+k_N\cdot e^{-r_N\cdot x+u_N})^{\sfrac{1}{k_N}}\bigg]^{-1},
\end{align*}
with the independent variable $x$ and where $\theta_N = (w_N,h_N,k_N,q_N,r_N,s_N,u_N)$ is the set of parameters. 
We then optimise $\theta_N$ with respect to the mean squared error between the fitting function and our enlarged table values for each sample size $N$ separately. 
We find that most of the parameters are approximately constant across sample sizes $N$. Thus, we set them to the mean value across all $N$ to reduce the number of parameters. Hence, it is sufficient to set $w_N,h_N,k_N,s_N$ and $u_N$ as constants and further set $q_N = r_N = b_N$ as the one remaining parameter depending on the sample size $N$. 
The resulting optimal values for $b_N$ are shown in Fig. \ref{fig:parameters_b} (teal stars). As they visually resemble a square root function of $N$, we model the parameter $b_N$ as the following function of $N$:
\begin{align*}
    b(N) =&  b_1\cdot \sqrt{N}+b_2.
\end{align*}
Note, that $b_1$ and $b_2$ are parameters independent of $N$. 
The final optimisation is to find $b_1$ and $b_2$ such that 
\begin{align*}
    \mathcal{E}(b_1,b_2) = \frac{1}{|S|L}\sum_{N\in \mathcal{S}}\sum_{i=1}^{L} (p_{Dip_i,N} - \hat{p}(Dip_i,\hat{b}(N)))^2
    \label{eq:MSE}
\end{align*}
is minimal. Here, $\mathcal{S}$ is the set of all sample sizes $N$, where $|\mathcal{S}|=63$ in the case of our enlarged table, and $L$ is the number of $(Dip,p)$-pairs in the extended table, here $L=307$. The optimal function within our optimisation scheme with respect to the mean squared error and with our bootstrapped table as data to be fitted is given by
\begin{align*}
    \hat{p}(x,\hat{b}(N)) = 1 &-
    \bigg[0.6\cdot(1+1.6 \cdot e^{-\hat{b}(N)\cdot x+6.5})^{\sfrac{1}{1.6}}\\
    &+~0.4\cdot(1+0.2\cdot e^{-\hat{b}(N)\cdot x+6.5})^{\sfrac{1}{0.2}}\bigg]^{-1},
\end{align*}
and 
\begin{align*}
\hat{b}(N) =  17.30784\cdot \sqrt{N}+12.04918
\end{align*}
where values for $\hat{b}_1 = 17.30784$, $\hat{b}_2 = 12.04918$ are rounded to five decimal places and correspond to the optimal estimators for $b_1$ and $b_2$.
The concatenated function $\hat{p}(x,\hat{b}(N))$ is smooth and well-defined for all $N \in \mathbb{N}^{+}$ and all $x \in (0,0.25]$. 

\textbf{Derivation:}
One of the benefits of our proposal is, that the fitted function is differentiable. This property is exploited in our later discussed subspace clustering algorithm. The \diptest can only be applied to a one-dimensional sample, which is why we always consider a projection axis $\rho$ for $d$-dimensional data sets $X$. Further, the data has to be sorted. Thus, the \dipval is returned for the sorted and projected data $\overline{X}_\rho$. From \cite{krause2005multimodal} we know that we can calculate the gradient vector of the \dipval on $\overline{X}_\rho$ with respect to the projection axis $\rho$. We will term this gradient $\nabla_\rho(Dip(\overline{X}_\rho))$. Details about the calculation of this gradient can be found in \cite{krause2005multimodal,skinnydip,dipext,dipencoder}. By using our differentiable sigmoid function to calculate the \dippvalue of the $Dip$, the \dippvalue is also continuously differentiable as a concatenation of differentiable functions. The gradient is the following, where we define $D := Dip(\overline{X}_\rho)$ and $b := \hat{b}(N)$ for easier readability:
\begin{flalign}
&\nabla_\rho(\hat{p}(D,b))= \big( -b\nabla_\rho(D) \big)e^{-bD+6.5} \cdot \nonumber\\ 
&\Big[ 0.6(1+1.6e^{-bD+6.5})^{\frac{1}{1.6}} + 0.4(1+0.2e^{-bD+6.5})^{\frac{1}{0.2}} \Big]^{-2} \cdot  \nonumber\\
& \Big[0.6(1+1.6e^{-bD+6.5})^{\frac{-0.6}{1.6}} + 0.4(1+0.2e^{-bD+6.5})^{\frac{0.8}{0.2}}\Big] \nonumber
\end{flalign}

\subsection{\Methoddip}
\label{sec:dipnsub}
We aim to show with a proof-of-concept that our differentiable \dippvalue function has great value for the data mining community. Therefore, we present the subspace clustering algorithm \Methoddip which is solely based on the \diptest. It is able to automatically define a lower-dimensional subspace and also to derive the number of clusters. For this, only a significance threshold is necessary, which indicates whether a distribution is unimodal. We first propose TailoredDip, a new extension of UniDip \cite{skinnydip}.

A problem with UniDip is that the tails of distributions are very generously identified as outliers. 
TailoredDip is able to better capture those tails. 
We achieve this by checking the spaces between the clusters for additional structures after the regular UniDip algorithm has terminated.
Therefore, we mirror the respective area between two clusters and calculate the \dippvalue. If this indicates multimodal structures, we identify appropriate modes and assign those points to the best fitting neighbouring cluster. 
Further, if outlier detection is not desired we use the following strategy to assign them either to the left or the right cluster: Instead of simply defining the mid point between neighbouring clusters as a decision boundary, we choose the point that corresponds to the intersection of the ECDF and the line between the right limit of the left cluster and the left limit of the right cluster. This handles different tails more accurately.
Details about TailoredDip as well as a pseudocode are given in the supplement (Sec. 1).

\begin{figure*}
\centering
\begin{subfigure}[t]{0.392\textwidth}
\includegraphics[width=\textwidth]{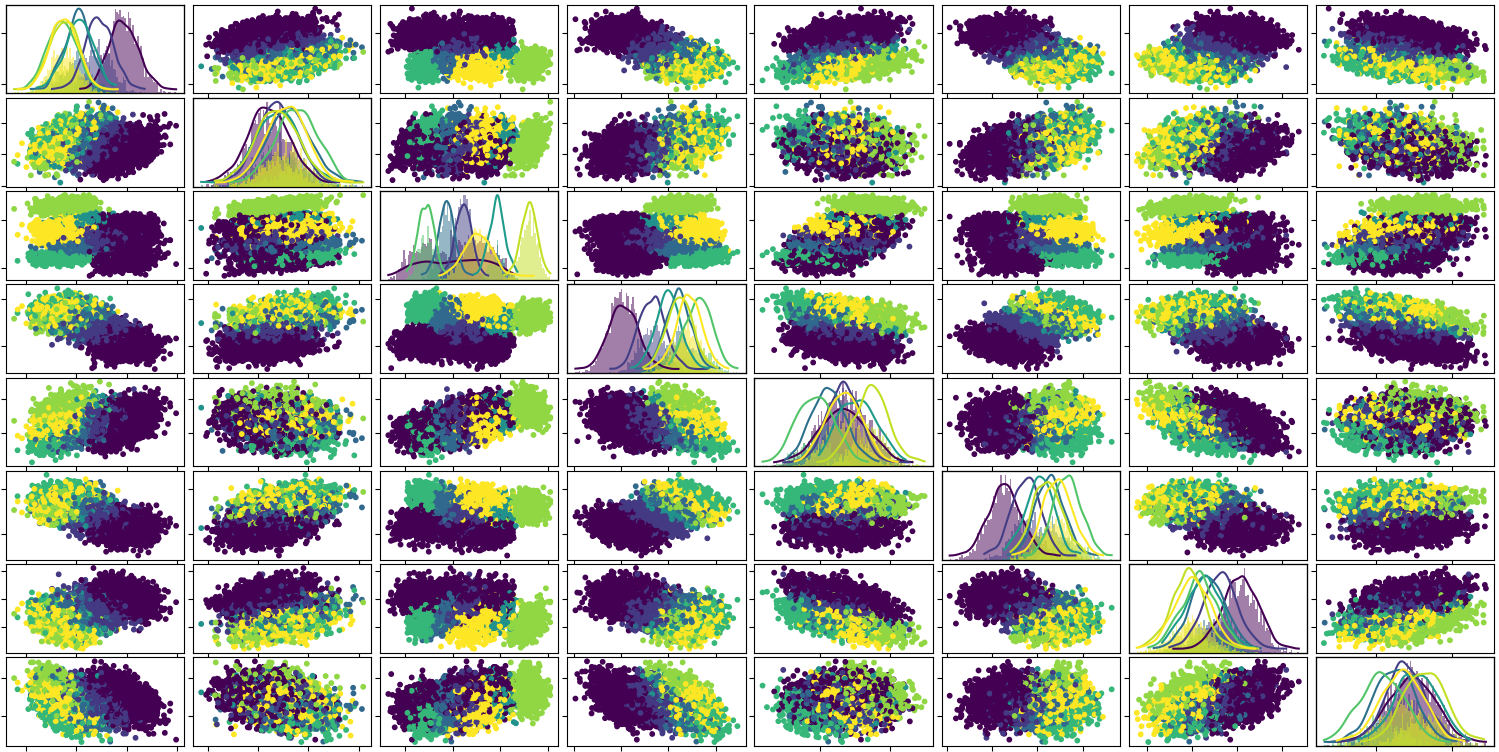}
\captionsetup{format=hang}
\subcaption{The input dataset.}
\label{fig:synth_input}
\end{subfigure}\hfill
\begin{subfigure}[t]{0.234\textwidth}
\includegraphics[width=\textwidth]{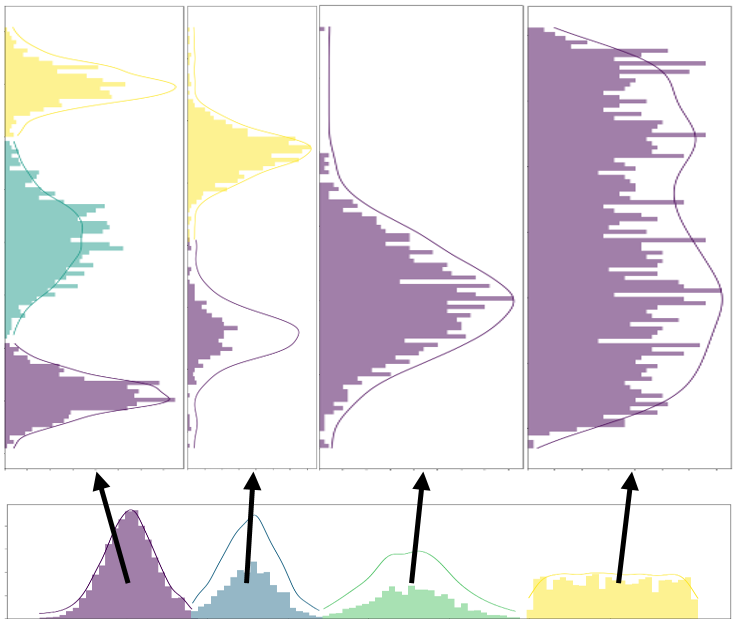}
\captionsetup{format=hang}
\subcaption{The two projections identified by \Methoddip.}
\label{fig:dipnsub_dimensions}
\end{subfigure}\hfill
\begin{subfigure}[t]{0.357\textwidth}
\includegraphics[width=\textwidth]{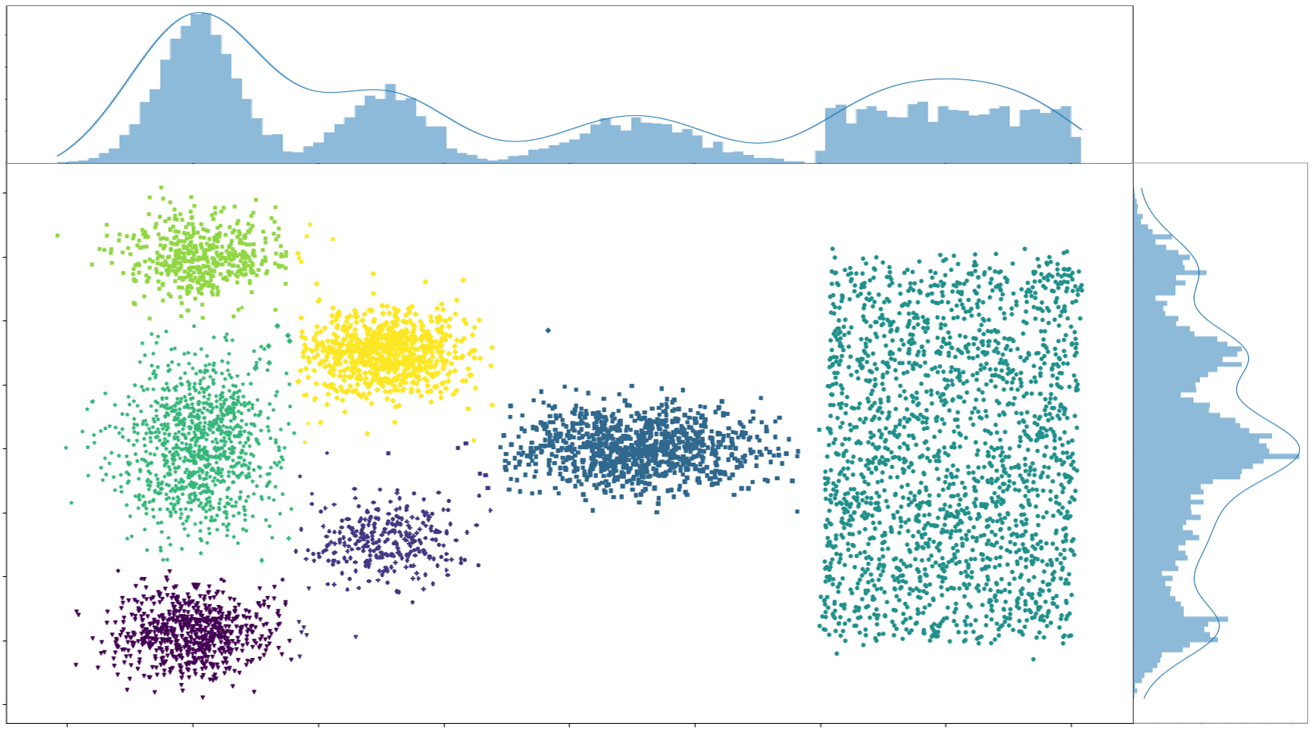}
\captionsetup{format=hang}
\subcaption{The final clustering result of \Methoddip.}
\label{fig:dipnsub_final}
\end{subfigure}
\caption{\textbf{(a)} Scatter matrix plot of an $8$-dimensional synthetic data set (colours correspond to ground-truth labels). \textbf{(b)} The horizontal histogram below illustrates the first projection identified by \Methoddip. The data is highly multimodal and therefore divided into $4$ clusters. \Methoddip now uses these cluster assignments to identify a second projection in which all clusters are as multimodal as possible. The second projection is shown vertically in the upper histograms, with respect to each existing cluster individually. It is easy to see that the first two clusters (purple and blue) are subdivided into $3$ and $2$ clusters, respectively. Thereafter, no multimodal third projection can be found. \textbf{(c)} The final clustering result of \Methoddip reveals a clear separation of the clusters.}
\label{fig:synth_example}
\end{figure*} 

Now that we can find clusters in one-dimensional samples, let us consider the multidimensional case. Here, we use the fact that SGD can be used to find a projection axis on which a data set shows a minimum \dippvalue. A naive approach would be to recursively select each cluster and, using the points of that cluster, find the projection axis on which those samples exhibit the greatest multimodality. The problem here is twofold. First, there is almost always some degree of multimodality in a set of objects, which means that one will identify a very high number of clusters. Second, the individual clusters are difficult to compare with each other, since each cluster forms its own subspace. Therefore, we want to successively identify those features in which as many objects as possible are contained in highly multimodal clusters. Thus, we recursively search for projection axes $\rho$ that minimise the following term:
\begin{equation}
    \frac{1}{N}\sum_{i=1}^{k}|C_i|\hat{p}\big(\text{Dip}(\overline{C_i^\rho}), \hat{b}(|C_i|)\big),
    \label{Eq:costFunction}
\end{equation}
where $C_i$ are the samples in cluster $i$ and $\overline{C_i^\rho}$ are the same samples projected to $\rho$ and sorted afterwards, i.e. $\overline{C_i^\rho} = sort\{\rho^Tc | c \in C_i\}$. To find these axes, we are inspired by \cite{dipext}. We start with the $q$ features that show the lowest sum of Dip-p-values weighted by their cluster sizes and use them as starting points for SGD with momentum. Further, we start at the first $q$ components of a PCA. \cite{dipext} has shown that $q=\log(d)$ is sufficient, where $d$ is the original number of features.
Using these $2q$ starting axes, we iteratively calculate the gradient with respect to all clusters. Here, we exploit that our fitted function enables us to directly calculate the gradient of Eq. \ref{Eq:costFunction}.
Having identified the best projection axis, we check whether more than $T\%$ of the objects are contained in clusters considered multimodal on this axis, where $T$ has to be set by the user. If this is the case, we apply TailoredDip to this axis. Each cluster is considered individually and divided into several
sub-clusters. The clusters thereby form hypercubes in the final feature space. 
Our method \Methoddip is presented in Algorithm \ref{alg:dipnsub} and an example of the subspace identification process is shown in Fig.~\ref{fig:synth_example}. 

\begin{algorithm2e}[t]
	\SetAlgoVlined
	\DontPrintSemicolon
	\KwIn{data set $X$, significance $\alpha$, threshold $T$}
	\KwOut{labels}
	\SetKw{KwWith}{with}
	$k=1; labels = [0, \dots, 0]; X_{fin} = []$\;
    \While{True}{
        $s = 1; \rho=\Vec{0}$\;
        $Q = \log(d)$ features with lowest weighted p-values $\cup$ first $\log(d)$ components of PCA\;
        \For{each $\rho_{tmp} \in$ Q}{
            Update $\rho_{tmp}$ with SGD using Eq. \ref{Eq:costFunction}\;
            $s_{tmp} = $ value of Eq. \ref{Eq:costFunction} using $\rho_{tmp}$\;
            \If{$s_{tmp} < s$}{
                $s = s_{tmp}; \rho = \rho_{tmp}$\;
            }
        }
        $P = \{$p-value(Dip($\overline{C_i^\rho}$), $|\overline{C_i^\rho}|$)$~|~i \in [1, k]\}$\;
        \uIf{$\frac{sum\{|C_i|~|~i \in [1, k] \wedge P_i < \alpha\}}{N} \ge T$}{
            \For{each cluster $i$ \KwWith $P_i < \alpha$}{
                $labels_{new}$ = TailoredDip($\overline{C_i^\rho}, \alpha$)\;
                update $labels$ using $labels_{new}$\;
            }
            $X_{fin} =$ combine $X_{fin}$ and $\{\rho^T x | x \in X\}$\;
            $X =$ keep features orthogonal to $\rho$\; 
        }\Else{break}
    }
	\Return{$labels, X_{fin}$}
	\caption{The \Methoddip algorithm \label{alg:dipnsub}}
\end{algorithm2e}

\section{Experiments and Results}
We will show the several benefits of our proposal in three main experimental sections. First, we show, that our calculated Dip-p-values are as reliable as the ones with the look-up table. Then we present runtime experiments that prove our method to be efficient, not only in `laboratory conditions', but also in practice when integrated in existing methods using Dip-p-values. Finally, we evaluate our subspace clustering algorithm \Methoddip to showcase the integration of the gradient of the \dippvalue in a practical data mining application. 

Our supplement, codes, enlarged ($Dip$, $p$)-pairs table and the used data sets are available at: \url{https://dx.doi.org/10.6084/m9.figshare.21916752}.

\begin{table*}[t]
\caption{Dip-p-values for different unimodal (\textbf{left}) and multimodal (\textbf{right}) distributions with varying sample sizes $N$.
All given values are averages for $100$ random samples $\pm$ standard deviation. Respective first, second and third rows per distribution show Dip-p-values calculated with methods `table' (T), `function' (F) and `bootstrapping' (B, $1000$ repetitions). Dip-p-values for multimodal distributions are multiplied by $100$; $\ast$: values obtained by $\sqrt{N}-$ interpolation, \textdagger: values not available.}
    \centering
      \resizebox{\textwidth}{!}{
    \begin{tabular}{l|l|cccc||l|cccc}
    \toprule
\textbf{unim. Distr.} & Method & $N=50$ &$N=234$ &$N=2345$&$N=100k$ & \textbf{multim. Distr.} & $N=50$ &$N=234$ &$N=2345$&$N=100k$\\
\midrule
\multirow{3}{*}{$\mathcal{N}(4,1)$}
& T & $0.77 \pm 0.24$& $0.86 \pm 0.19^*$ & $0.97 \pm 0.07^*$ & \textdagger & $\mathcal{N}(4,1)$  & $8.94 \pm 15.0$ & $0.06 \pm 0.23^*$  & $0.00 \pm 0.00^*$ & \textdagger\\
& F & $0.77 \pm 0.24$& $0.86 \pm 0.19$ &  $0.97 \pm 0.07$ & $1.00 \pm 0.02$ & $\cup$             & $8.83 \pm 14.9$ & $0.09 \pm 0.25$    & $0.00 \pm 0.00$ & $0.00 \pm 0.00$   \\  
& B & $0.77 \pm 0.24$& $0.86 \pm 0.19$ &  $0.97 \pm 0.07$ & $1.00 \pm 0.02$ & $\mathcal{N}(0,1)$ & $8.78 \pm 15.0$ & $0.06 \pm 0.21$    & $0.00 \pm 0.00$ & $0.00 \pm 0.00$  \\\midrule
\multirow{3}{*}{$\mathcal{T}_{nc}(4,2,0,1)$}
& T& $ 0.80 \pm 0.21$  & $0.89 \pm 0.14^*$ & $0.98 \pm 0.03^*$ & \textdagger & $\mathcal{T}_{nc}(4,2,0,1)$     & $0.79 \pm 2.20$ & $0.00 \pm 0.00^*$  & $0.00 \pm 0.00^*$ & \textdagger\\
& F & $ 0.80 \pm 0.21$ & $0.90 \pm 0.14$ & $0.98 \pm 0.03$ & $1.00 \pm 0.00$ & $\cup$                      & $0.83 \pm 2.07$ & $0.00 \pm 0.00$    & $0.00 \pm 0.00$ & $0.00 \pm 0.00$ \\ 
& B& $ 0.80 \pm 0.21$  & $0.90 \pm 0.14$ & $0.99 \pm 0.03$ & $1.00 \pm 0.00$ & $\mathcal{T}_{nc}(4,2,7,1)$ & $0.74 \pm 2.17$ & $0.00 \pm 0.00$    & $0.00 \pm 0.00$ & $0.00 \pm 0.00$ \\ \midrule
\multirow{3}{*}{$\mathcal{L}(0,2)$}
& T& $0.85 \pm 0.19$  & $0.95 \pm 0.11^*$ & $0.99 \pm 0.04^*$ & \textdagger   & $\mathcal{L}(0,2)$  & $24.1 \pm 24.9$ & $2.36 \pm 7.87^*$  & $0.00 \pm 0.00^*$ & \textdagger\\
& F& $0.85 \pm 0.19$  & $0.95 \pm 0.11$   & $0.99 \pm 0.04$ & $1.00 \pm 0.00$ & $\cup$              & $23.9 \pm 25.1$ & $2.37 \pm 7.82$    & $0.00 \pm 0.00$ & $0.00 \pm 0.00$\\
& B& $0.85 \pm 0.19$  & $0.95 \pm 0.11$   & $0.99 \pm 0.04$ & $1.00 \pm 0.00$ & $\mathcal{L}(7,2)$  & $23.8 \pm 24.8$ & $2.34 \pm 8.06$    & $0.00 \pm 0.00$ & $0.00 \pm 0.00$ \\\midrule

\end{tabular}
}
    \label{tab:consistent_p}
\end{table*}

\subsection{Reliable Computation}
\label{sec:reliability_consistency}
One important advantage of our fitted function is, that it provides Dip-p-values for all sample sizes $N$. Table \ref{tab:consistent_p} shows that our `function' method is consistent with the look-up `table' and `bootstrap' methods as we produce basically the same Dip-p-values, not only for different unimodal distributions ($\mathcal{N}(4,1)$ = normal distribution with centre $4$ and variance $1$; $\mathcal{T}_{nc}(4,2,0,1)$ = non central student's t-distribution with $4$ degrees of freedom, non-centrality parameter $2$, centre $0$ and scaling $1$; $\mathcal{L}(0,2)$ = Laplace distribution with centre $0$ and scaling $2$), but also for multimodal distributions, which we create by combining samples of the same unimodal distribution, but with a different centre. In the supplement (Sec. 2.3) we show tables with a total of $23$ distributions, where we observe the same behaviour. 

Comparing to the enlarged bootstrapped table described in Sec. \ref{ssec:table_ext} our function performs better than the `table' method, with the quality of both methods being measured in mean squared error (MSE).
In this case, bootstrapping can be seen as some kind of ground truth, however it is not practical as runtime is a major issue, especially for large $N$. We will see this in more detail in the next section. 
Regarding the `table' method, errors for sample sizes greater than $72,000$ had to be ignored for the calculation of this MSE, because it cannot provide any Dip-p-values in that case. 
While we achieve an MSE of $3.43\cdot 10^{-6}$ for all $N$, the `table' results in an MSE of $7.92 \cdot 10^{-6}$. To check how well our function generalises with respect to $N$, we calculate the MSE for a set of $N$ chosen as the mean values between each two $N$ of our enlarged table. Those $62$ values were not used for our function fit. The result is an MSE of $3.14\cdot 10^{-6}$ for `function' and $8.12\cdot 10^{-6}$ for `table'. Hence, we outperform the `table' method in both cases.

\subsection{Computing Time}
\label{sec:runtime}
\begin{figure}[t]
\centering
\includegraphics[width=\linewidth]{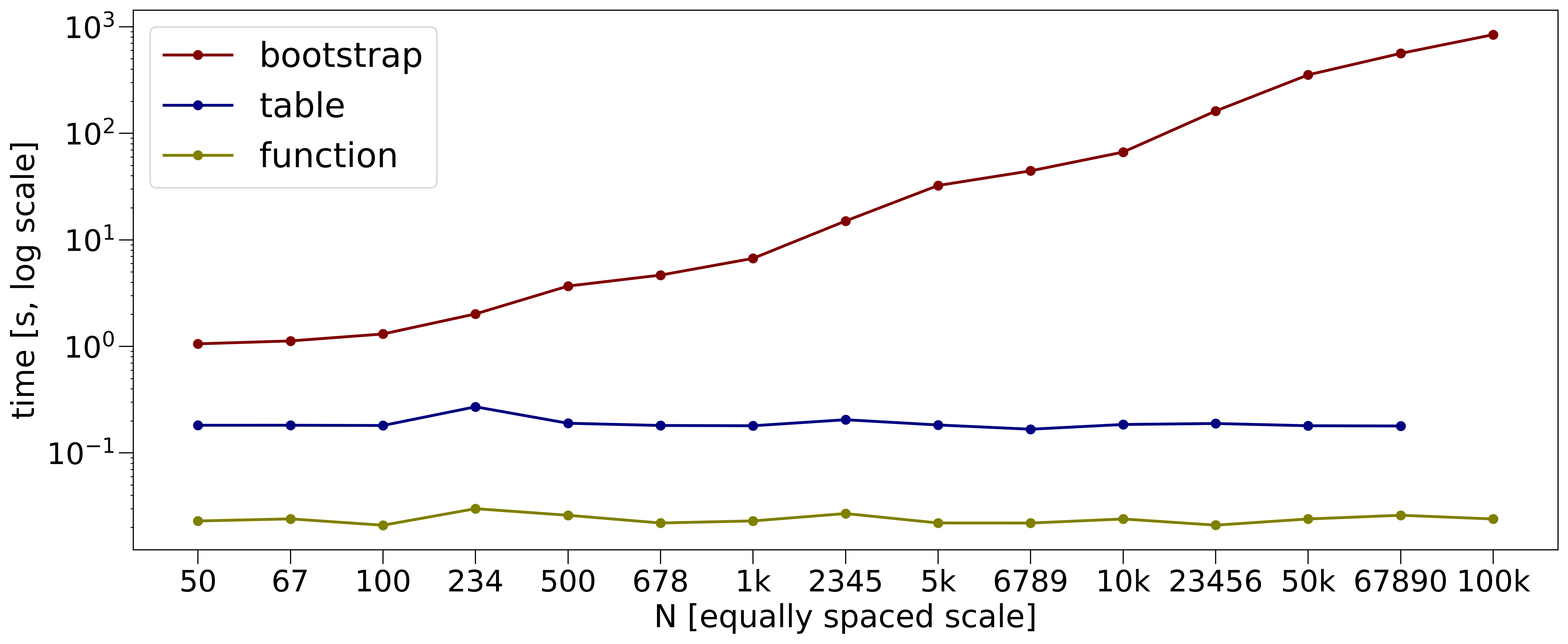}

\caption{Runtime in seconds [s] on a logarithmic scale for the calculations of $100$ Dip-p-values per sample size with the three methods `bootstrap', `table' and `function' summed up over all $23$ distribution scenarios as described in the supplement (Sec. $2.3$). 
}
\label{fig:runtime1}
\end{figure}

In the first experiment, we sum up the runtimes of all the \dippvalue calculations for the $23$ distribution cases shown in the supplement (Table \ref{tab:consistent_p} is a selection of six of them). 
These are shown in Fig. \ref{fig:runtime1}. Note, that the y-scale is logarithmic. 
Note also the missing value for $N = 100k$ for the `table' method as here the calculation of Dip-p-values is not possible. We can observe, that the calculation of Dip-p-values is fastest with our function, although it should be noted that the speed-up relative to the `table' method is only marginal and could be due to implementation details. 
The differences concerning the calculation time become of practical value, when the number of Dip-p-values to be calculated gets larger. In Table \ref{tab:runtime2} we can see how the algorithms DipMeans, projected DipMeans and SkinnyDip have decreasing runtime, when our method is used instead of the look-up table or bootstrapping. Information about the data sets are given in the supplement (Sec. 2.2). As expected, our function method does not degrade the clustering performance. Across all three algorithms and all data sets, the normalised mutual information (NMI)
remains stable with an average difference between `table' and `function' of $9.00\cdot 10^{-3}$ and $1.17\cdot 10^{-2}$ between `function' and `bootstrap'. 
Runtime is comparable to the `table' method and improves notably compared to `bootstrapping' as we save 50\%, 92\% and 99\% for DipMeans, projected DipMeans and SkinnyDip, respectively.
\begin{table*}[t]
\caption{Average NMI and runtime (RT - in seconds) results for DipMeans, p. DipMeans and SkinnyDip using the \dippvalue calculation methods `table' (T), `function' (F) and `bootstrap' (B) after $10$ runs.}
  \centering
  \resizebox{\textwidth}{!}{
  \begin{tabular}{l||c|c|c|c|c|c||c|c|c|c|c|c||c|c|c|c|c|c}
    \toprule
    \multirow{3}{*}{Dataset} & \multicolumn{6}{c||}{DipMeans} & \multicolumn{6}{c||}{p. DipMeans} & \multicolumn{6}{c}{SkinnyDip} \\
    \cmidrule{2-19}
    & \multicolumn{3}{c|}{NMI} & \multicolumn{3}{c||}{RT} &  \multicolumn{3}{c|}{NMI} & \multicolumn{3}{c||}{RT} &  \multicolumn{3}{c|}{NMI} & \multicolumn{3}{c}{RT} \\
     & T & F & B & T & F & B & T & F & B & T & F & B & T & F & B & T & F & B \\
     \midrule
    SYNTH & 0.64 & 0.64  & 0.64 & 6.87 & 5.82 & 8.92 & 0.86 & 0.86 & 0.85 & 0.76 & 0.74 & 9.03 & 0.16 & 0.16 & 0.16 & 0.02 & 0.01 & 14.66\\
    BANK & 0.31 & 0.31 & 0.30 & 13.00 & 6.93 & 43.71 & 0.30 & 0.31 & 0.30 & 7.70 & 5.46 & 69.81 & 0.13 & 0.13 & 0.13 & 0.01 & 0.00 & 2.56\\
    USER & 0.00 & 0.00 & 0.00 & 0.05 & 0.02 & 0.08 & 0.34 & 0.34 & 0.35 & 1.54 & 1.13 & 11.37 & 0.15 & 0.15 & 0.15 & 0.00 & 0.00 & 0.68\\
    HTRU2 & 0.00 & 0.00  & 0.00 & 28.87 & 28.32 & 29.03 & 0.17 & 0.17 & 0.17 & 2.20 & 2.18 & 19.05 & 0.08 & 0.08 & 0.08 & 0.04 & 0.04 & 40.78\\
    ALOI & 0.96 & 0.96 & 0.92 & 0.43 & 0.22 & 0.84 & 0.49 & 0.51 & 0.49 & 9.03 & 6.68 & 54.36 & 0.17 & 0.15 & 0.17 & 0.06 & 0.04 & 3.68\\
    MICE & 0.00 & 0.00 & 0.00 & 0.19 & 0.12 & 0.27 & 0.53 & 0.54 & 0.53 & 91.47 & 50.21  & 607.21 & 0.00 & 0.00 & 0.00 & 0.01 & 0.01 & 3.76\\
    AIBO & 0.00 & 0.00 & 0.00 & 0.09 & 0.04 & 0.14 & 0.28 & 0.33 & 0.27 & 20.02 & 5.33 & 129.60 & 0.02 & 0.02 & 0.02 & 0.02 & 0.01 & 3.29\\
    MOTE & 0.35 & 0.35 & 0.35 & 0.85 & 0.62 & 1.21 & 0.00 & 0.00 & 0.00 & 0.03 & 0.02 & 0.37 & 0.00 & 0.00 & 0.00 & 0.03 & 0.02 & 8.73\\
    SYMB & 0.82 & 0.82 & 0.82 & 1.50 & 1.15 & 2.23 & 0.70 & 0.74 & 0.70 & 36.70 & 5.64  & 73.69 & 0.02 & 0.02 & 0.02 & 0.11 & 0.10 & 5.16\\
    OLIVE & 0.50 & 0.50 & 0.50 & 0.06 & 0.04 & 0.16 & 0.64 & 0.52 & 0.64 & 1.26 & 0.15 & 3.08 & 0.04 & 0.04 & 0.04 & 0.03 & 0.02 & 0.39\\
    \bottomrule
  \end{tabular}}
  \label{tab:runtime2}
\end{table*}

\begin{table*}[htbp]
\caption{Maximum NMI results of different common subspace and Dip-based $k$-estimation algorithms after $10$ runs. The resulting number of clusters and dimensions is given in brackets. Best result in bold, runner-up dotted. ($k=$ number of clusters, $d$ = data set dimensionality, KM = k-means, $\dagger$: no results due to non-trivial errors).}
\centering
\resizebox{\textwidth}{!}{
\begin{tabular}{l|l|llllll|lll}
\toprule
\multirow{2}{*}{Dataset ($k/d$)} &  \multicolumn{7}{c}{Common Subspace Algorithms} & \multicolumn{3}{|c}{Dip-based $k$-estimation Algorithms}\\
\cmidrule{2-11}
 & Dip'n'Sub                  & PCA+KM               & ICA+KM  & DipExt+KM & LDA-KM      & SubKM & FOSSCLU    & DipMeans  & p. DipMeans & SkinnyDip \\ 
\midrule 
SYNTH (7/8)   & \textbf{0.97}  (7/2)        & 0.87 (7/4)         & 0.38 (7/7) & 0.69 (7/1) & 0.88 (7/6)   & 0.87 (7/6) & 0.90 (7/5)  & 0.64 (3/8) & $\dotuline{0.92}$ (6/8)  & 0.16 (2/8) \\ 
BANK  (2/4)   & $\dotuline{0.41}$ (7/3)                  & 0.03 (2/2)         & 0.01 (2/2) & \textbf{0.83} (2/1) & 0.01 (2/1)   & 0.03 (2/1) & 0.01 (2/4) & 0.32 (41/4)  & 0.32 (36/4) & 0.13 (2/4) \\ 
USER  (4/5)   & $\dotuline{0.52}$ (10/1)         & 0.43 (4/5)         & 0.03 (4/4)  & 0.40 (4/2) & 0.49 (4/3)   & 0.46 (4/3) & \textbf{0.65} (4/2) & 0.00 (1/5) & 0.36 (33/5)  & 0.15 (2/5)\\ 
HTRU2 (2/8)   & \textbf{0.38} (3/2)         & 0.03 (2/2)         & 0.30 (2/2) & 0.03 (2/1) & 0.28 (2/1)   & 0.03 (2/1) & $\dotuline{0.32}$ (2/5)  & 0.00 (1/66) & 0.18 (7/66) & 0.08 (1/66) \\ 
ALOI  (4/66)  & $\dotuline{0.98}$ (4/1)                  & \textbf{1.00} (4/35)  & \textbf{1.00} (4/4)  &  \textbf{1.00} (4/5) & \textbf{1.00} (4/3)  & \textbf{1.00} (4/3) & $\dagger$ & 0.96 (5/66) & 0.52 (69/66)  & 0.17 (16/66) \\
MICE  (8/68)  & $\dotuline{0.55}$ (6/3)                  & 0.27 (8/6)         & 0.39 (8/8) & \textbf{0.59} (8/3) & $\dagger$   & 0.29 (8/7) & 0.33 (8/5)  & 0.00 (1/68) & 0.54 (278/68) & 0.00 (1/68)\\ 
AIBO (2/70) & \textbf{0.68} (2/1)  & \textbf{0.68} (2/20)  & 0.56 (2/2) & $\dotuline{0.60}$ (2/3)  & 0.30 (2/1) & \textbf{0.68} (2/1) & 0.52 (2/5) & 0.00 (1/70) & 0.34 (35/70) & 0.02 (3/70) \\
MOTE  (2/84)  & $\dotuline{0.37}$ (2/1)                  & 0.30 (2/42)        & $\dotuline{0.37}$ (2/2) & \textbf{0.41} (2/2) & 0.28 (2/1)   & 0.35 (2/1) & 0.08 (2/5) & 0.36 (3/84) & 0.00 (1/84) & 0.00 (1/84)\\ 
SYMB  (6/398) & \textbf{0.84} (5/3)         & 0.80 (6/6)         & 0.79 (6/6) & 0.65 (6/2) & 0.80 (6/5)   & 0.80 (6/5) & $\dagger$  & $\dotuline{0.82}$ (5/398) & 0.74 (16/398) & 0.02 (2/398)\\ 
OLIVE (4/570) & 0.57 (4/2)                  & 0.68 (4/4)  & 0.69 (4/4) & $\dotuline{0.73}$ (4/58)  & $\dagger$ & \textbf{0.75} (4/3) & 0.16 (4/3) & 0.50 (2/570) & 0.68 (9/570) & 0.04 (1/570) \\
\bottomrule
\end{tabular}}
\label{tab:dipnsub}
\end{table*}

\subsection{\Methoddip Evaluation}
\label{sec:subspace_res}
We evaluate our algorithm \Methoddip and competitors in terms of clustering performance using the normalised mutual information (NMI). This score attains values between $0$ and $1$, where $0$ indicates a purely random label assignment and a value close to $1$ is a perfect clustering result.

\textbf{Comparison Methods:}
We compare to multiple approaches that define a common subspace for all clusters. This includes dimensionality reduction methods like PCA, ICA or DipExt, which we combine with k-means, and the algorithms LDA-k-means, FOSSCLU and SubKmeans. Furthermore, since \Methoddip is able to estimate the number of clusters, we compare with the Dip-based $k$-estimation methods DipMeans, projected DipMeans and SkinnyDip.

For PCA, we set the number of components such that 90\% of the variance is preserved, and for ICA, it equals $k$. The significance level is set to $0.01$ for all Dip-based methods. The range in which FOSSCLU can determine the number of subspace dimensions with MDL is defined as $[1, 5]$. All other parameters were set as described in the respective papers.
Regarding \Methoddip we set $T=0.15$ and the significance to $0.01$. For SGD, we choose a momentum of $0.95$ and a step-size of $0.1$ (for USER, ALOI, AIBO, SYMB, OLIVE) or $0.01$ (for SYNTH, BANK, HTRU2, MICE, MOTE). Since none of the above data sets contains outliers, we assign all points to the best matching cluster by using the strategy described in Sec. \ref{sec:dipnsub}.

\textbf{Quantitative Analyses:}
Table \ref{tab:dipnsub} shows the results of our algorithm \Methoddip and our competitors on a wide range of data sets (see supplement Sec. 2.2 for details).
Note, that compared to the other subspace algorithms we do not know the ground truth number of clusters.
Nevertheless, we are competitive compared to subspace and Dip-based $k$-estimation methods as we rank first $4$ times and second $5$ times in terms of NMI.
On ALOI, we are slightly inferior, but we only need a single feature for our result.
Overall, \Methoddip achieves a good ratio of NMI to the identified number of cluster-relevant features. Our method identifies rather small subspaces (maximum number of features is $3$ for BANK, MICE and SYMB), which in combination with the good NMI values suggests that those features are particularly relevant for clustering. 
This can be interesting for a visual evaluation of the results especially in the unsupervised domain. 
Furthermore, our estimation of $k$ is notably better than that of other Dip-based methods (which we outperform every time with regard to NMI except for OLIVE). While those procedures identify the correct number of clusters only once, we manage it in 50\% of the cases. Especially projected DipMeans seems to heavily overestimate the number of clusters.
This confirms our hypothesis that in many cases one can detect additional multimodal structures when looking at a single cluster. Therefore, we benefit from considering only those projection axes relevant to all clusters.

\section{Conclusion}
In this paper, we propose a differentiable function to translate Dip-values to Dip-p-values. This provides an automatic and fast translation for any desired sample size. We show that our method is effective as our Dip-p-values show lower squared errors than previously used look-up tables. Further, it is efficient regarding computing time. Finally, we underpin its practical relevance by integrating our function in the subspace clustering algorithm \Methoddip. Here, we show how our proposal enables the use of gradient descent for the Dip-test's $p$-value. Future efforts may attempt to integrate those ideas into deep learning applications.

\bibliographystyle{siam}
\bibliography{bibliography}

\end{document}


\title{\Large Supplement to `Extension of the Dip-test Repertoire - Efficient and Differentiable p-value Calculation for Clustering'}

\author{Lena G. M. Bauer\thanks{Authors contributed equally.}~$^,$\footnote{
Faculty of Computer Science, ds:UniVie, University of Vienna, Vienna, Austria. \{lena.bauer, christian.boehm, claudia.plant\}@univie.ac.at}~$^,$\footnote{UniVie Doctoral School Computer Science.}
\and Collin Leiber$^{\ast,}$\thanks{LMU Munich \& MCML, Munich, Germany. leiber@dbs.ifi.lmu.de}     
\and Christian Böhm$^\dagger$
\and Claudia Plant$^\dagger$
}

\date{}

\maketitle

\fancyfoot[R]{\scriptsize{Copyright \textcopyright\ 2023 by SIAM\\
Unauthorized reproduction of this article is prohibited}}

\section{TailoredDip}
In the following we explain TailoredDip, which adds two extensions to the UniDip \cite{skinnydip} algorithm. First, we show how a cluster can be expanded to include the tails of a distribution and then how outliers can be assigned to an appropriate cluster.

\subsection{Capturing the Tails}
\begin{figure}[th]
\centering
\begin{subfigure}[t]{0.49\linewidth}
\includegraphics[width=\linewidth]{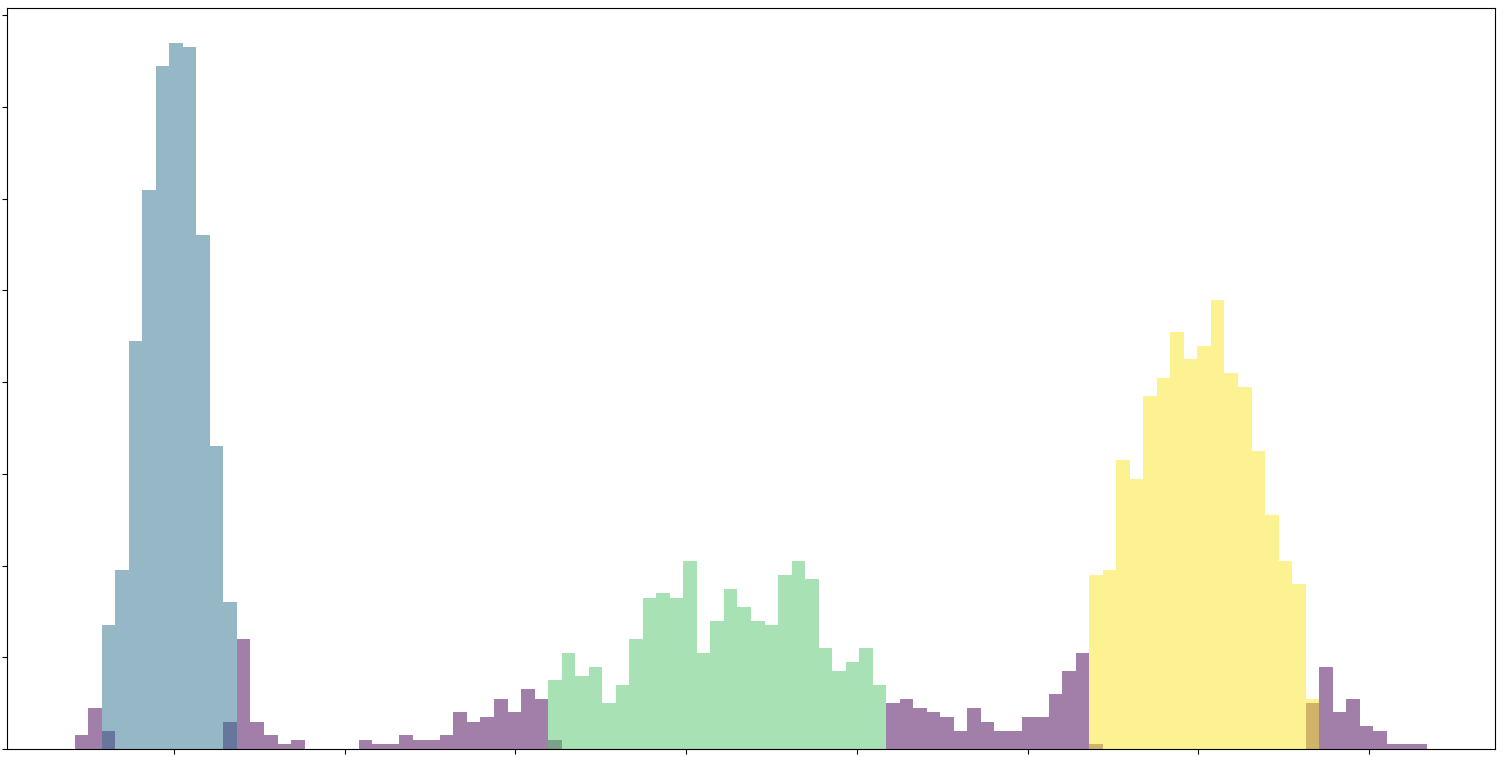}
\captionsetup{format=hang}
\subcaption{Result of UniDip\\NMI = $0.81$.}
\label{fig:unidip}
\end{subfigure}
\begin{subfigure}[t]{0.49\linewidth}
\includegraphics[width=\linewidth]{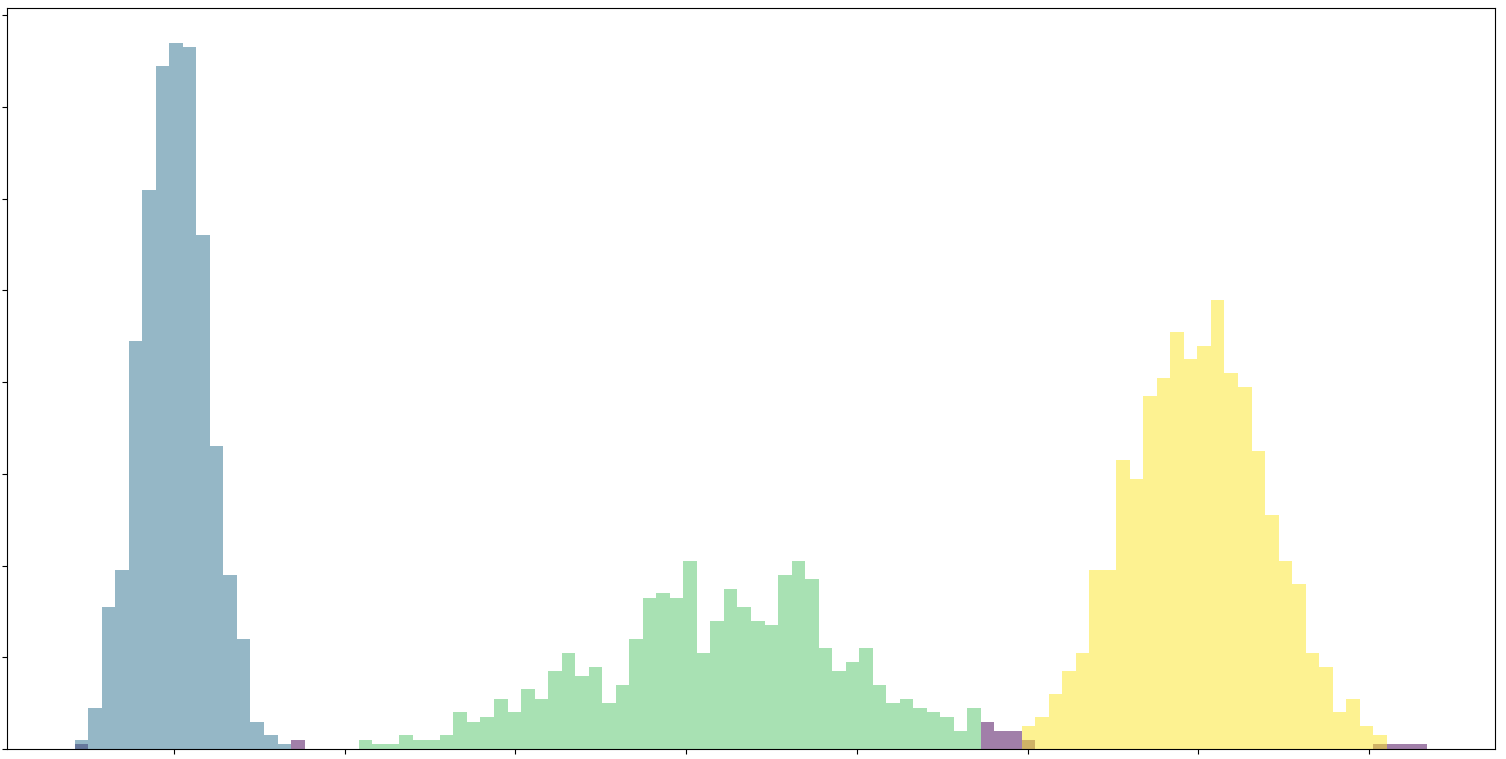}
\captionsetup{format=hang}
\subcaption{Result of TailoredDip\\NMI = $0.96$.}
\label{fig:unidip_plus}
\end{subfigure}
\begin{subfigure}[t]{0.49\linewidth}
\includegraphics[width=\linewidth]{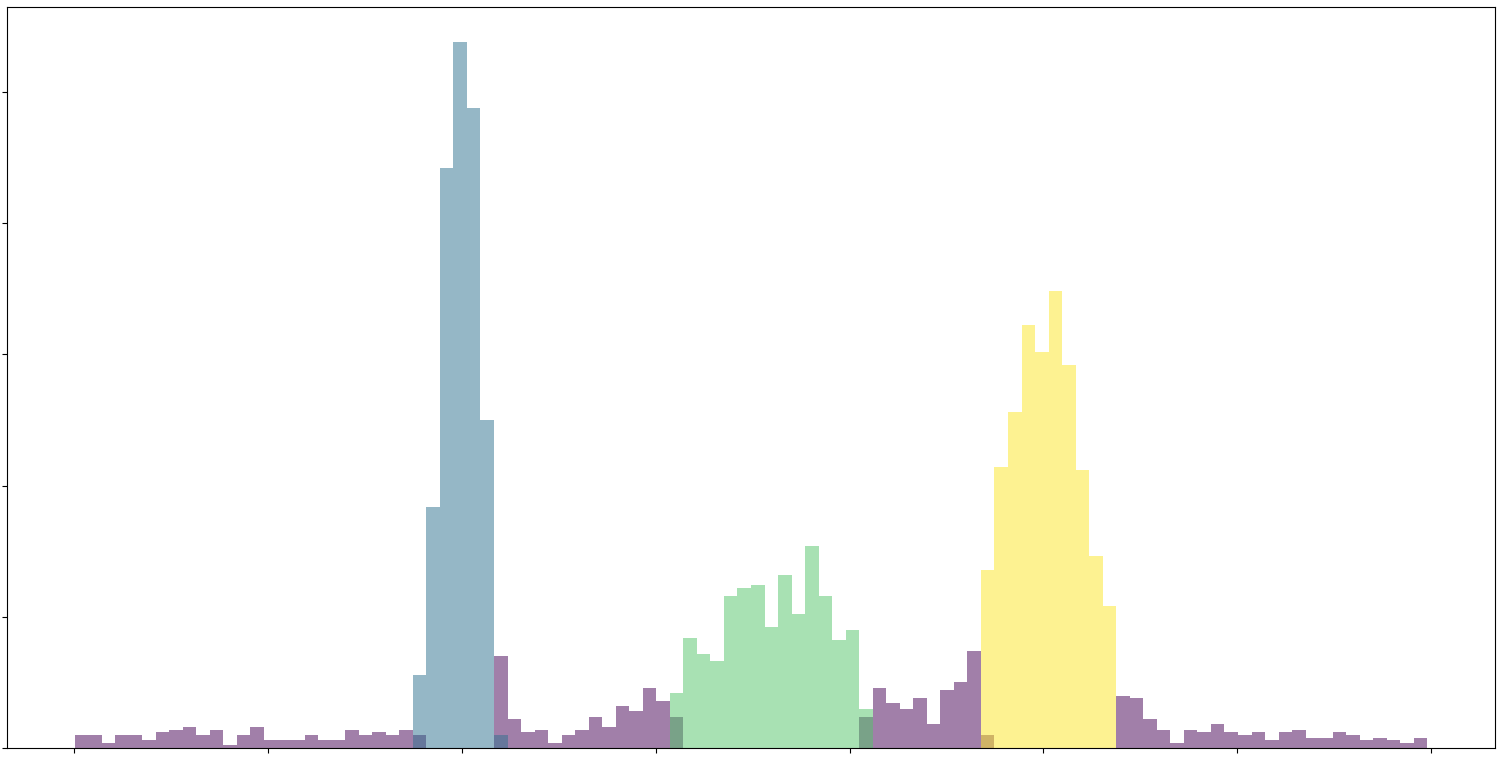}
\captionsetup{format=hang}
\subcaption{Result of UniDip\\NMI = $0.70$ (noisy data).}
\label{fig:unidip_noise}
\end{subfigure}
\begin{subfigure}[t]{0.49\linewidth}
\includegraphics[width=\linewidth]{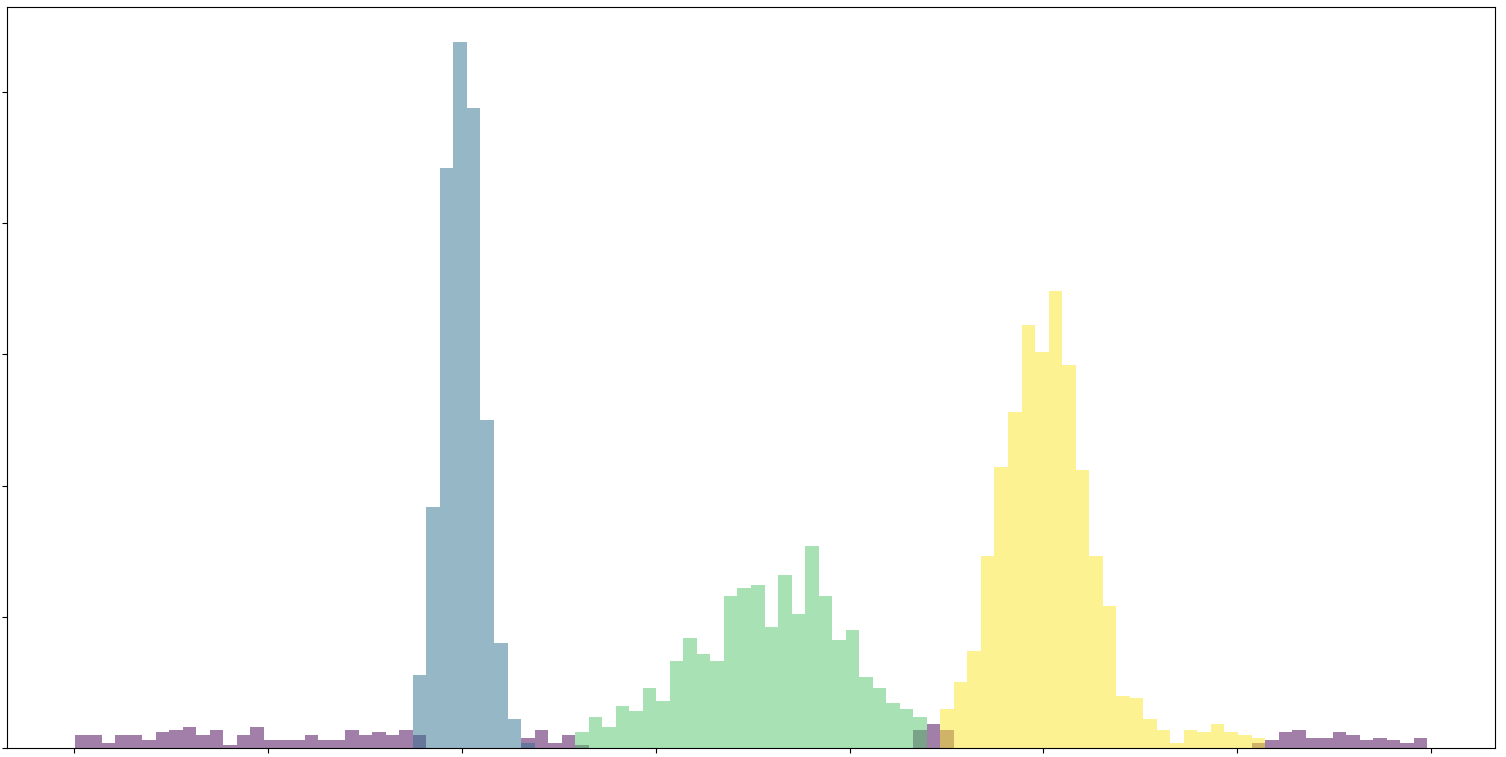}
\captionsetup{format=hang}
\subcaption{Result of TailoredDip\\NMI = $0.78$ (noisy data).}
\label{fig:unidip_plus_noise}
\end{subfigure}
\caption{Results of TailoredDip and UniDip on a sample data set consisting of three Gaussian clusters. The identified clusters are coloured in blue, green, and yellow respectively. Outliers are shown in purple.
}
\end{figure}

As mentioned in the paper, UniDip has problems correctly identifying the tails of distributions.
These are usually labeled as noise.
This behavior can be observed in Fig. \ref{fig:unidip}. The Gaussian clusters are not completely captured, but only the densest parts of the distributions. 
The same applies when uniformly distributed noise is added to the data (see Fig. \ref{fig:unidip_noise}). TailoredDip is superior in capturing the tails of the distributions in both cases. This is also confirmed by the normalised mutual information (NMI) score and can be seen in Fig. \ref{fig:unidip_plus} and \ref{fig:unidip_plus_noise}. We achieve this improvement by checking the spaces between the clusters for additional structures after the regular UniDip algorithm has terminated. Although these structures are no longer significant enough to be regarded as independent clusters by UniDip, they can still be part of a cluster. Therefore, we mirror the respective area between two clusters and calculate the \dippvalue. If this indicates that there are still multimodal structures left, we again search for appropriate modes.
In order to check whether a found structure matches the adjacent clusters, we apply a strategy that has been described in \cite{dipdeck}. Here, the closest $2|S|$ samples of the respective cluster combined with the newly found structure $S$ are used to calculate the \dippvalue. If this value indicates unimodality, the structure will be added to that cluster and the process is repeated.
The described procedure is shown in Algorithm \ref{alg:unidipplus}. Since in our case a lot of Dip-p-values have to be calculated, a fast calculation of Dip-p-values is favourable.

\subsection{Assigning Noise}
\begin{figure}[t]
\centering
\includegraphics[width=\linewidth]{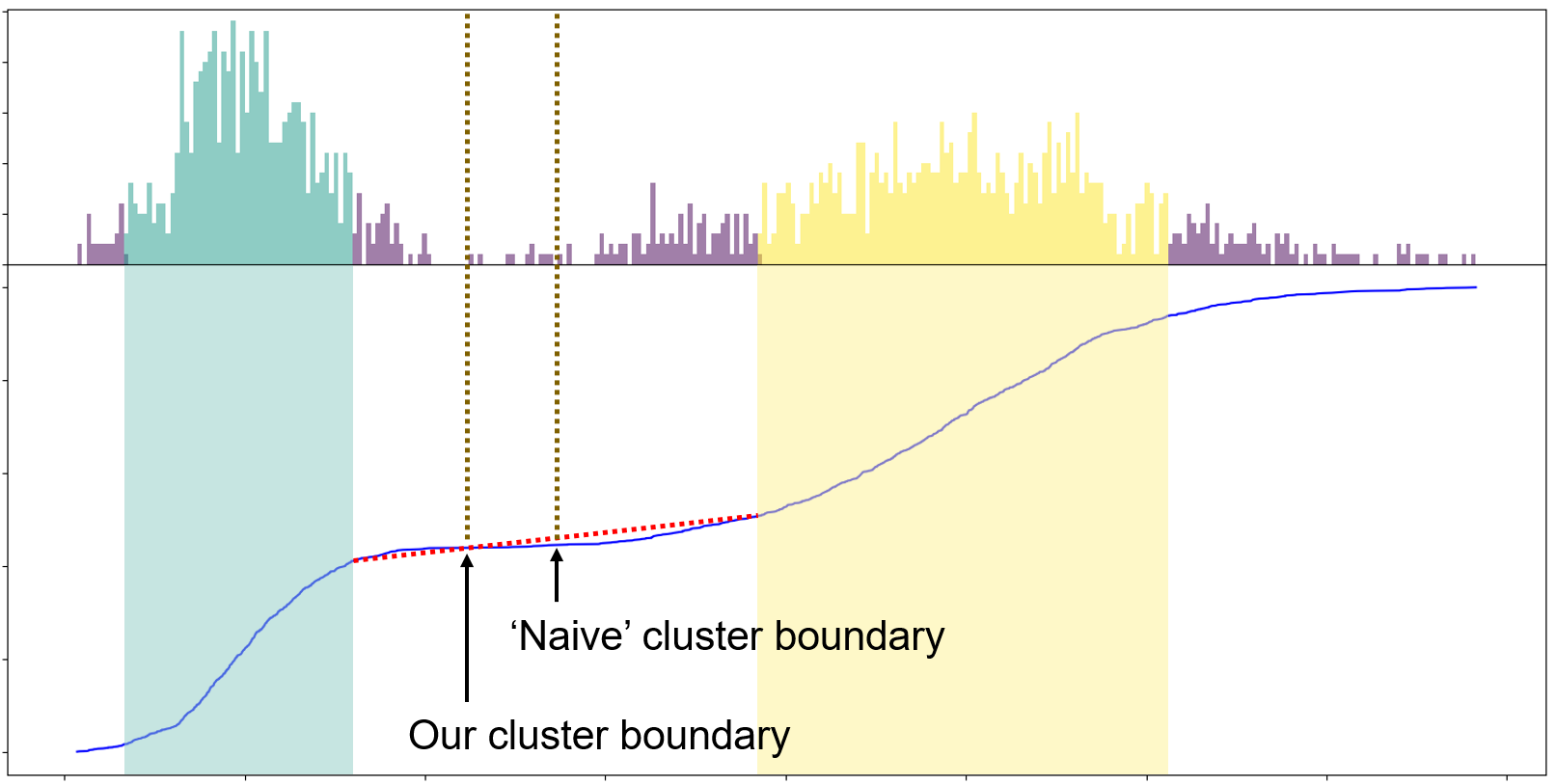}
\caption{Visualisation of our strategy to assign outliers to the neighbouring clusters. [Top] A histogram of the data. The left (teal) cluster originates from a $\mathcal{N}(0,1)$ distribution with $700$ samples, the right (yellow) cluster originates from a $\mathcal{N}(10,2.5)$ distribution with $900$ samples. The outliers are shown in purple. [Bottom] The ECDF of the data is shown in blue. The areas of the clusters are highlighted in their respective colours. The dotted red line indicates the connection line between the end of the teal and the beginning of the yellow cluster. The right vertical brown line marks the position of a naive cluster boundary, which corresponds to the centre of the red line. Our boundary corresponds to the intersection of the red line with the ECDF and captures the different tails of the two cluster much better.}
\label{fig:assign_outliers}
\end{figure}

We also present a strategy for assigning outliers to clusters, paying attention to the different tails of the surrounding distributions. In terms of one-dimensional data, it makes sense to define a threshold between every two clusters, indicating whether an outlier is more likely to belong to the left or right cluster. A naive approach would now be to simply set the midpoint between the end of the left and the start of the right cluster. This strategy was chosen in \cite{dipencoder}, for example. However, this approach completely ignores the existing structures, since it is irrelevant whether a cluster ends abruptly (e.g. in case of an uniform distribution) or fades out slowly (e.g. in case of a normal distribution). To pay attention to these properties, we consider the Empirical Cumulative Distribution Function (ECDF), which is also used to calculate the \dipval.
Here, we draw a straight line from the last point of the left cluster to the first point of the right cluster. In Fig. \ref{fig:assign_outliers} this is represented by the dotted red line.
We now define the intersection of this line with the ECDF of the data as the cluster boundary.
Looking at this point in Fig. \ref{fig:assign_outliers} (left vertical line) we can see that it separates the tails of the two distributions much better than the naive strategy (right vertical line) since it better captures the higher standard deviation of the right cluster. If more than one intersection occurs, we choose the one closest to the midpoint between the clusters.

\begin{algorithm2e}[t!]
	\SetAlgoVlined
	\DontPrintSemicolon
	\KwIn{one-dimensional data set $X$, significance $\alpha$}
	\KwOut{labels}
	\SetKw{KwGoTo}{go to}
	// Get initial clusters by running UniDip\;
    $labels, k$ = UniDip($X, \alpha$)\;
    \For{$i=0;~ i \le k;~ i \mathrel{+}= 1$}{
        \uIf{$i==0$}{\label{"lnl:start"}
            $X_\text{sub}$ = samples left of first cluster\;
        }\uElseIf{$i==k$}{
            $X_\text{sub}$ = samples right of last cluster\;
        }\Else{
            $X_\text{sub}$ = samples between cluster $i$ and $i+1$\;
        }
        // Is $X_\text{sub}$ uniformly distributed (only noise)?\;
        $X_\text{mirror}$ = mirror $X_\text{sub}$\;
        $p = $p-value(Dip($X_\text{mirror}$), $|X_\text{mirror}|$)\;
        \If{$p<\alpha$}{
            $labels_\text{new}, k_\text{new} = $UniDip($X_\text{sub}, \alpha$)\;
            $X_\text{first}$ = combine cluster $i$ with the first new cluster // \textit{(ignore if $i==0$)}\;
            $p_\text{first}$ = p-value(Dip($X_\text{first}$), $|X_\text{first}|$)\;
            $X_\text{last}$ = combine cluster $i+1$ with the last new cluster // \textit{(ignore if $i==k$)}\;
            $p_\text{last}$ = p-value(Dip($X_\text{last}$), $|X_\text{last}|$)\;
            \uIf{$i\neq 0$ and $p_\text{first} \ge \alpha$ and ($k_\text{new}\neq1$ or $p_\text{first} \ge p_\text{last}$)}{
                Update $labels$ by adding all entries with $labels_\text{new} == 1$ to cluster $i$\;
            }\uElseIf{$i\neq k$ and $p_\text{last} \ge \alpha$ and ($k_\text{new}\neq1$ or $p_\text{last} > p_\text{first}$)}{
                Update $labels$ by adding all entries with $labels_\text{new} == k_\text{new}$ to cluster $i+1$\;
            }
        }
        \If{Cluster $i$ or $i+1$ was updated}{\KwGoTo line \ref{"lnl:start"}}
    }
	\Return{$labels$}
	\caption{The TailoredDip algorithm \label{alg:unidipplus}}
\end{algorithm2e}

\section{Additional Information for the Experiments}

\subsection{Computational Setup}

\Methoddip as well as the algorithms DipMeans \cite{dipmeans}, projected DipMeans \cite{p_dipmeans}, SkinnyDip \cite{skinnydip}, DipExt \cite{dipext}, LDA-k-means \cite{ldakmeans} and SubKmeans \cite{subkmeans} are all implemented in Python. Regarding FOSSCLU \cite{fossclu} we use the Java implementation as referenced in the paper.
We conduct all runtime experiments on a machine with an Intel Core i7-5600U CPU with 2.60GHz and 8GB RAM. Further, we use Python 3.7 and in case of FOSSCLU, we use Java 8 due to compatibility issues.

\subsection{Data Sets}

We conduct experiments on $9$ real world data sets and one synthetic data set (the latter can be seen in the main paper in Fig. 4). Banknotes (BANK), User Knowledge (USER), HTRU2 and Mice Protein (MICE) are numerical data sets from the UCI repository\footnote{\url{https://archive.ics.uci.edu}}. SonyAIBO (AIBO), MoteStrain (MOTE), Symbols (SYMB) and OliveOil (OLIVE) are time series data sets\footnote{\url{https://www.timeseriesclassification.com}}, 
and ALOI\footnote{\url{https://aloi.science.uva.nl/}} is an image data collection. ALOI was preprocessed as described in \cite{isaac}, resulting in $288$ samples devided into $4$ clusters. Other than ALOI, no data set did receive any pre-processing, except that features with a variance of $0$ were removed. Note, that TailoredDip only works with continuous features, otherwise each value can be recognized as a separate mode. A summary of the data sets is given in Table \ref{tab:datasummary}.

\begin{table}[t]
\centering
\caption{Summary of the data sets ($N$ = number of data points, $d$ = dimensionality, $k$ = number of clusters).}
\begin{tabular}{lrrr}
\toprule
Dataset & $N$   & $d$   & $k$  \\ 
\midrule
SYNTH \quad & 6,300      & 8                & 7 \\ 
BANK \quad  & 1,372      & 4                & 2 \\
USER \quad  & 403        & 5                & 4 \\
HTRU2 \quad & 17,898     & 8                & 2 \\
ALOI \quad  & 288        & 66               & 4 \\
MICE \quad  & 1,077      & 68               & 8 \\
AIBO \quad  & 621        & 70               & 2 \\
MOTE \quad  & 1,272      & 84               & 2 \\
SYMB \quad  & 1,020      & 398              & 6 \\
OLIVE \quad & 60         & 570              & 4 \\
\bottomrule
\end{tabular}
\label{tab:datasummary}
\end{table}

\subsection{Interpolate Look-up Table}
We would like to briefly explain how missing values in the state-of-the-art look-up table are interpolated. Basically two interpolations must be performed. First, the values for the number of samples that lie below and above the input $N$ must be searched for in the table.
By using these two values we are able to interpolate all relevant ($Dip$,$p$)-pairs in relation to $\sqrt{N}$. In this interpolated array we search for the Dip-values that are below and above our input $Dip$ to interpolate the \dippvalue linearly.

\subsection{Large Distribution Table}
In Tables \ref{tab:unim_distr} and \ref{tab:multi_distr} we show Dip-p-value calculations with the three methods `table' (T), `function' (F) and `bootstrap' (B) for samples of $15$ different sample sizes and a total of $23$ distribution scenarios. In all cases, we can observe that our fitted function produces basically the same Dip-p-values as the other two methods. For this evaluation we first consider $8$ different unimodal distributions: 
\begin{itemize}
    \item $\mathcal{N}(a,b) \dots$ normal distribution with mean $a$ and standard deviation $b$
    \item $\mathcal{T}(d,a,b) \dots$ students-t distribution with $d$ degrees of freedom, centre $a$ and scaling $b$
    \item $\mathcal{L}(a,b) \dots$ Laplace distribution with centre $a$ and scaling $b$
    \item $\mathcal{U}(a,b) \dots$ uniform distribution between $a$ and $b$
    \item $\mathcal{G}(s,a,b) \dots$ Gamma distribution with shape parameter $s$, centre $a$ and scaling $b$
    \item $\mathcal{E}(a,b) \dots$ exponential distribution with centre $a$ and scaling $b$
    \item $\mathcal{B}(s,r,a,b) \dots$ Beta distribution with shape parameters $s$ and $r$, centre $a$ and scaling $b$
    \item $\mathcal{T}_{nc}(d,c,a,b) \dots$ non central students-t distribution with $d$ degrees of freedom, non centrality $c$, centre $a$ and scaling $b$
\end{itemize}

First, Table \ref{tab:unim_distr} shows results for these distributions, with only the listed distributions involved individually.
We then generate $8$ multimodal distributions by combining $\frac{N}{2}$ samples from one distribution with $\frac{N}{2}$ samples from the same distribution with a different centre. Additionally, we consider $7$ cases, where we generate samples by choosing half the points from $\mathcal{N}(4,1)$ and the other half from one of the other unimodal distributions. These combinations always show multimodal structure. An exception is the case of samples from $\mathcal{N}(4,1) \cup \mathcal{T}_{nc}(4,2,0,1)$. We include this combination to have a relatively unambiguous case between unimodal and multimodal. Our function performs reliably in all cases as can be seen in Table \ref{tab:multi_distr}.

\bibliographystyle{siam}
\bibliography{bibliography}

\begin{landscape}

\begin{table}[]
\caption{Dip-p-values for different unimodal distributions with varying sample sizes $N$.
All given values are averages for $100$ random samples $\pm$ standard deviation. Respective first, second and third rows per distribution show Dip-p-values calculated with methods `table' (T), `function' (F) and `bootstrapping' (B, $1000$ repetitions); $\ast$: values obtained by $\sqrt{N}-$ interpolation, \textdagger: values not available.}
\resizebox{1.36\textwidth}{!}{
\begin{tabular}{l|l|ccccccccccccccc}
\toprule
Distr.                        & \multicolumn{1}{c|}{Meth.} & \multicolumn{1}{c}{$N=50$} & \multicolumn{1}{c}{$N=67$} & \multicolumn{1}{c}{$N=100$} & $N=234$         & $N=500$        & $N=678$         & $N=1k$         & $N=2345$        & $N=5k$         & $N=6789$        & $N=10k$        & $N=23456$       & $N=50k$        & $N=67890$       & $N=100k$       \\ \midrule
\multirow{3}{*}{$\mathcal{N}(4,1)$}       & T              &$ 0.77 \pm 0.24              $&$ 0.81 \pm 0.21^*            $&$ 0.80 \pm 0.20              $&$  0.86 \pm 0.19^* $&$ 0.91 \pm 0.14 $&$ 0.94 \pm 0.13^*  $&$ 0.94 \pm 0.09 $&$ 0.97 \pm 0.07^*  $&$  0.99 \pm 0.03 $&$ 0.98 \pm 0.05^* $&$ 0.99 \pm 0.04  $&$  0.99 \pm 0.01^*  $&$  1.00 \pm 0.02  $&$  1.00 \pm 0.01^*  $&  \textdagger   \\
                              & F                          &$  0.77 \pm 0.24              $&$  0.81 \pm 0.21              $&$  0.81 \pm 0.20               $&$  0.86 \pm 0.19   $&$  0.91 \pm 0.14  $&$  0.94 \pm 0.13   $&$  0.95 \pm 0.09  $&$  0.97 \pm 0.07   $&$  0.99 \pm 0.03  $&$  0.98 \pm 0.05   $&$  0.99 \pm 0.04  $&$  1.00 \pm 0.01   $&$  1.00 \pm 0.02  $&$  1.00 \pm 0.01   $&$  1.00 \pm 0.02$ \\
                               &  B                        &$ 0.77 \pm 0.24              $&$  0.81 \pm 0.21              $&$  0.81 \pm 0.20               $&$  0.86 \pm 0.19   $&$  0.91 \pm 0.14  $&$  0.94 \pm 0.13   $&$  0.95 \pm 0.09  $&$  0.97 \pm 0.07   $&$  0.99 \pm 0.03  $&$  0.98 \pm 0.05   $&$  0.99 \pm 0.04  $&$  1.00 \pm 0.01   $&$  1.00 \pm 0.02  $&$  1.00 \pm 0.01   $&$  1.00 \pm 0.02$ \\ \midrule
\multirow{3}{*}{$\mathcal{T}(4,0,1)$}     & T              &$ 0.78 \pm 0.22              $&$  0.85 \pm 0.19^*             $&$  0.84 \pm 0.20               $&$  0.88 \pm 0.18^*  $&$  0.94 \pm 0.10  $&$  0.95 \pm 0.09^*  $&$  0.97 \pm 0.07  $&$  0.97 \pm 0.07^*  $&$  0.99 \pm 0.04  $&$  0.99 \pm 0.03^*  $&$  0.99 \pm 0.02  $&$  1.00 \pm 0.01^*  $&$  1.00 \pm 0.00  $&$  1.00 \pm 0.01^*  $&  \textdagger   \\
                              & F                          &$ 0.78 \pm 0.22              $&$  0.85 \pm 0.19              $&$  0.85 \pm 0.20               $&$  0.88 \pm 0.18   $&$  0.94 \pm 0.10  $&$  0.96 \pm 0.09   $&$  0.97 \pm 0.07  $&$  0.97 \pm 0.07   $&$  0.99 \pm 0.04  $&$  0.99 \pm 0.03   $&$  0.99 \pm 0.02  $&$  1.00 \pm 0.01   $&$  1.00 \pm 0.00  $&$  1.00 \pm 0.01   $&$  1.00 \pm 0.00$ \\
                              & B                          &$ 0.78 \pm 0.22              $&$  0.85 \pm 0.19              $&$  0.85 \pm 0.20               $&$  0.88 \pm 0.18   $&$  0.94 \pm 0.10  $&$  0.96 \pm 0.09   $&$  0.97 \pm 0.07  $&$  0.97 \pm 0.07   $&$  0.99 \pm 0.04  $&$  0.99 \pm 0.02   $&$  0.99 \pm 0.02  $&$  1.00 \pm 0.01   $&$  1.00 \pm 0.00  $&$  1.00 \pm 0.01   $&$  1.00 \pm 0.00$ \\ \midrule
\multirow{3}{*}{$\mathcal{L}(0,2)$}       & T              &$ 0.85 \pm 0.19              $&$  0.88 \pm 0.18^*             $&$  0.92 \pm 0.11               $&$  0.95 \pm 0.11^*  $&$  0.98 \pm 0.04  $&$  0.98 \pm 0.04^*  $&$  0.99 \pm 0.01  $&$  0.99 \pm 0.04^*  $&$  1.00 \pm 0.00  $&$  1.00 \pm 0.00^*  $&$  1.00 \pm 0.00  $&$  1.00 \pm 0.00^*  $&$  1.00 \pm 0.00  $&$  1.00 \pm 0.00^* $& \textdagger   \\
                              & F                          &$ 0.85 \pm 0.19              $&$  0.89 \pm 0.18              $&$  0.92 \pm 0.11               $&$  0.95 \pm 0.11   $&$  0.98 \pm 0.04  $&$  0.99 \pm 0.04   $&$  0.99 \pm 0.01  $&$  0.99 \pm 0.04   $&$  1.00 \pm 0.00  $&$  1.00 \pm 0.00   $&$  1.00 \pm 0.00  $&$  1.00 \pm 0.00   $&$  1.00 \pm 0.00  $&$  1.00 \pm 0.00   $&$  1.00 \pm 0.00$ \\
                              & B                          &$ 0.85 \pm 0.19              $&$  0.89 \pm 0.18              $&$  0.92 \pm 0.11               $&$  0.95 \pm 0.11   $&$  0.98 \pm 0.04  $&$  0.99 \pm 0.03   $&$  1.00 \pm 0.01  $&$  0.99 \pm 0.04   $&$  1.00 \pm 0.00  $&$  1.00 \pm 0.00   $&$  1.00 \pm 0.00  $&$  1.00 \pm 0.00   $&$  1.00 \pm 0.00  $&$  1.00 \pm 0.00   $&$  1.00 \pm 0.00$ \\ \midrule
\multirow{3}{*}{$\mathcal{U}(0,2)$}       & T              &$ 0.52 \pm 0.29              $&$  0.50 \pm 0.28^*             $&$  0.46 \pm 0.28               $&$  0.53 \pm 0.29^*  $&$  0.52 \pm 0.28  $&$  0.54 \pm 0.31^*  $&$  0.48 \pm 0.30  $&$  0.50 \pm 0.28^*  $&$  0.56 \pm 0.28  $&$  0.54 \pm 0.29^*  $&$  0.52 \pm 0.32  $&$  0.50 \pm 0.30^*  $&$  0.53 \pm 0.28  $&$  0.51 \pm 0.30^* $& \textdagger   \\
                              & F                          &$ 0.52 \pm 0.30              $&$  0.51 \pm 0.28              $&$  0.46 \pm 0.28               $&$  0.53 \pm 0.30   $&$  0.52 \pm 0.29  $&$  0.54 \pm 0.31   $&$  0.48 \pm 0.30  $&$  0.49 \pm 0.29   $&$  0.56 \pm 0.28  $&$  0.54 \pm 0.30   $&$  0.52 \pm 0.32  $&$  0.50 \pm 0.30   $&$  0.53 \pm 0.28  $&$  0.51 \pm 0.30   $&$  0.56 \pm 0.28$ \\
                              & B                          &$ 0.52 \pm 0.29              $&$  0.51 \pm 0.28              $&$  0.46 \pm 0.28               $&$  0.53 \pm 0.30   $&$  0.52 \pm 0.29  $&$  0.54 \pm 0.31   $&$  0.48 \pm 0.30  $&$  0.50 \pm 0.29   $&$  0.56 \pm 0.28  $&$  0.53 \pm 0.30   $&$  0.52 \pm 0.32  $&$  0.50 \pm 0.30   $&$  0.53 \pm 0.28  $&$  0.51 \pm 0.31   $&$  0.56 \pm 0.28$ \\ \midrule
\multirow{3}{*}{$\mathcal{G}(2,-1,1)$}    & T              &$ 0.76 \pm 0.23              $&$  0.78 \pm 0.22^*             $&$  0.81 \pm 0.21               $&$  0.88 \pm 0.14^*  $&$  0.91 \pm 0.14  $&$  0.91 \pm 0.11^*  $&$  0.93 \pm 0.13  $&$  0.97 \pm 0.08^*  $&$  0.98 \pm 0.05  $&$  0.98 \pm 0.05^*  $&$  0.99 \pm 0.02  $&$  0.99 \pm 0.02^*  $&$  1.00 \pm 0.01  $&$  1.00 \pm 0.00^* $& \textdagger   \\
                              & F                          &$ 0.76 \pm 0.24              $&$  0.79 \pm 0.22              $&$  0.81 \pm 0.21               $&$  0.88 \pm 0.14   $&$  0.91 \pm 0.14  $&$  0.92 \pm 0.11   $&$  0.94 \pm 0.13  $&$  0.97 \pm 0.08   $&$  0.99 \pm 0.05  $&$  0.99 \pm 0.05   $&$  0.99 \pm 0.02  $&$  1.00 \pm 0.02   $&$  1.00 \pm 0.01  $&$  1.00 \pm 0.00   $&$  1.00 \pm 0.00$ \\
                              & B                          &$ 0.76 \pm 0.23              $&$  0.79 \pm 0.22              $&$  0.81 \pm 0.21               $&$  0.88 \pm 0.14   $&$  0.91 \pm 0.14  $&$  0.92 \pm 0.11   $&$  0.94 \pm 0.13  $&$  0.98 \pm 0.08   $&$  0.99 \pm 0.05  $&$  0.99 \pm 0.05   $&$  0.99 \pm 0.02  $&$  1.00 \pm 0.01   $&$  1.00 \pm 0.00  $&$  1.00 \pm 0.00   $&$  1.00 \pm 0.00 $\\ \midrule
\multirow{3}{*}{$\mathcal{E}(0,1)$}       & T              &$ 0.72 \pm 0.26              $&$  0.79 \pm 0.21^*             $&$  0.79 \pm 0.23               $&$  0.86 \pm 0.18^*  $&$  0.92 \pm 0.11  $&$  0.95 \pm 0.10^*  $&$  0.94 \pm 0.11  $&$  0.98 \pm 0.05^*  $&$  0.99 \pm 0.03  $&$  0.99 \pm 0.02^*  $&$  1.00 \pm 0.01  $&$  1.00 \pm 0.00^*  $&$  1.00 \pm 0.00  $&$  1.00 \pm 0.00^*  $&  \textdagger   \\
                              & F                          &$ 0.72 \pm 0.26              $&$  0.79 \pm 0.21              $&$  0.80 \pm 0.23               $&$  0.86 \pm 0.18   $&$  0.93 \pm 0.11  $&$  0.95 \pm 0.10   $&$  0.94 \pm 0.11  $&$  0.98 \pm 0.05   $&$  0.99 \pm 0.04  $&$  1.00 \pm 0.02   $&$  1.00 \pm 0.01  $&$  1.00 \pm 0.00   $&$  1.00 \pm 0.00  $&$  1.00 \pm 0.00   $&$  1.00 \pm 0.00$ \\
                              & B                          &$ 0.72 \pm 0.26              $&$  0.79 \pm 0.21              $&$  0.79 \pm 0.23               $&$  0.86 \pm 0.18   $&$  0.93 \pm 0.11  $&$  0.95 \pm 0.10   $&$  0.94 \pm 0.11  $&$  0.98 \pm 0.05   $&$  0.99 \pm 0.04  $&$  1.00 \pm 0.02   $&$  1.00 \pm 0.01  $&$  1.00 \pm 0.00   $&$  1.00 \pm 0.00  $&$  1.00 \pm 0.00   $&$  1.00 \pm 0.00$ \\ \midrule
\multirow{3}{*}{$\mathcal{B}(2,2,1,1)$}   & T              &$ 0.65 \pm 0.25              $&$  0.66 \pm 0.25^*             $&$  0.73 \pm 0.25               $&$  0.81 \pm 0.18^*  $&$  0.82 \pm 0.22  $&$  0.84 \pm 0.21^*  $&$  0.84 \pm 0.21  $&$  0.90 \pm 0.16^*  $&$  0.96 \pm 0.07  $&$  0.96 \pm 0.09^*  $&$  0.98 \pm 0.05  $&$  0.98 \pm 0.05^*  $&$  0.99 \pm 0.02  $&$  0.99 \pm 0.01^*  $&  \textdagger   \\
                              & F                          &$ 0.65 \pm 0.25              $&$  0.67 \pm 0.26              $&$  0.73 \pm 0.26               $&$  0.82 \pm 0.18   $&$  0.82 \pm 0.22  $&$  0.84 \pm 0.21   $&$  0.85 \pm 0.21  $&$  0.90 \pm 0.16   $&$  0.96 \pm 0.07  $&$  0.96 \pm 0.09   $&$  0.98 \pm 0.05  $&$  0.99 \pm 0.05   $&$  0.99 \pm 0.02  $&$  1.00 \pm 0.01   $&$  0.99 \pm 0.02$ \\
                              & B                          &$ 0.65 \pm 0.26              $&$  0.66 \pm 0.25              $&$  0.73 \pm 0.25               $&$  0.81 \pm 0.19   $&$  0.82 \pm 0.22  $&$  0.84 \pm 0.21   $&$  0.84 \pm 0.21  $&$  0.90 \pm 0.16   $&$  0.96 \pm 0.07  $&$  0.97 \pm 0.09   $&$  0.98 \pm 0.05  $&$  0.99 \pm 0.05   $&$  0.99 \pm 0.02  $&$  1.00 \pm 0.01   $&$  1.00 \pm 0.02$ \\ \midrule
\multirow{3}{*}{$\mathcal{T}_{nc}(4,2,0,1)$} & T           & $ 0.80 \pm 0.21              $&$  0.78 \pm 0.23^*             $&$  0.85 \pm 0.17             $&$  0.89 \pm 0.14^*  $&$  0.96 \pm 0.06  $&$  0.94 \pm 0.10^*  $&$  0.95 \pm 0.08  $&$  0.98 \pm 0.03^*  $&$  0.99 \pm 0.04  $&$  0.99 \pm 0.02^*  $&$  0.99 \pm 0.01  $&$  1.00 \pm 0.01^*  $&$  1.00 \pm 0.00  $&$  1.00 \pm 0.00^* $& \textdagger   \\
                              & F                          & $0.80 \pm 0.21              $&$  0.78 \pm 0.23              $&$  0.85 \pm 0.17               $&$  0.90 \pm 0.14   $&$  0.96 \pm 0.06  $&$  0.95 \pm 0.10   $&$  0.96 \pm 0.08  $&$  0.98 \pm 0.03   $&$  0.99 \pm 0.04  $&$  0.99 \pm 0.02   $&$  1.00 \pm 0.01  $&$  1.00 \pm 0.01   $&$  1.00 \pm 0.00  $&$  1.00 \pm 0.00   $&$  1.00 \pm 0.00$ \\
                              & B                          & $0.80 \pm 0.21              $&$  0.78 \pm 0.23              $&$  0.85 \pm 0.17               $&$  0.90 \pm 0.14   $&$  0.96 \pm 0.06  $&$  0.95 \pm 0.10   $&$  0.96 \pm 0.08  $&$  0.99 \pm 0.03   $&$  0.99 \pm 0.04  $&$  0.99 \pm 0.02   $&$  1.00 \pm 0.01  $&$  1.00 \pm 0.01   $&$  1.00 \pm 0.00  $&$  1.00 \pm 0.00   $&$  1.00 \pm 0.00$ \\ \bottomrule
\end{tabular}
}
\label{tab:unim_distr}
\end{table}

\end{landscape}

\begin{landscape}

\begin{table}[]
\caption{Dip-p-values for different combinations of distributions with varying sample sizes $N$.
All given values are averages for $100$ random samples $\pm$ standard deviation. Respective first, second and third rows per distribution show Dip-p-values calculated with methods `table' (T), `function' (F) and `bootstrapping' (B, $1000$ repetitions). Values are multiplied by $100$. $\ast$: values obtained by $\sqrt{N}-$ interpolation, \textdagger: values not available.}
\resizebox{1.36\textwidth}{!}{
\begin{tabular}{l|l|ccccccccccccccc}
\toprule
Distr.       & \multicolumn{1}{c|}{Meth.} & \multicolumn{1}{c}{$N=50$} & \multicolumn{1}{c}{$N=67$} & \multicolumn{1}{c}{$N=100$} & $N=234$         & $N=500$        & $N=678$         & $N=1k$         & $N=2345$        & $N=5k$         & $N=6789    $    & $N=10k$        & $N=23456$       & $N=50k$        & $N=67890$       & $N=100k$       \\ \midrule
$\mathcal{N}(4,1)$       & T    &$ 8.94 \pm 15.0             $&$ 6.97 \pm 14.1^*            $&$ 4.95 \pm 10.4              $&$ 0.06 \pm 0.23^* $&$ 0.00 \pm 0.00 $&$ 0.00 \pm 0.00^* $&$ 0.00 \pm 0.00 $&$ 0.00 \pm 0.00^* $&$ 0.00 \pm 0.00 $&$ 0.00 \pm 0.00^* $&$ 0.00 \pm 0.00 $&$ 0.00 \pm 0.00^* $&$ 0.00 \pm 0.00 $&$ 0.00 \pm 0.00^* $& \textdagger   \\
$\cup$       &$ F               $&$ 8.83 \pm 14.9             $&$ 7.03 \pm 14.2             $&$ 4.87 \pm 10.3              $&$ 0.09 \pm 0.25  $&$ 0.00 \pm 0.00 $&$ 0.00 \pm 0.00  $&$ 0.00 \pm 0.00 $&$ 0.00 \pm 0.00  $&$ 0.00 \pm 0.00 $&$ 0.00 \pm 0.00  $&$ 0.00 \pm 0.00 $&$ 0.00 \pm 0.00  $&$ 0.00 \pm 0.00 $&$ 0.00 \pm 0.00  $&$ 0.00 \pm 0.00$ \\
$\mathcal{N}(0,1)$       & B    &$ 8.78 \pm 15.0             $&$ 6.81 \pm 13.9             $&$ 4.97 \pm 10.5              $&$ 0.06 \pm 0.21  $&$ 0.00 \pm 0.00 $&$ 0.00 \pm 0.00  $&$ 0.00 \pm 0.00 $&$ 0.00 \pm 0.00  $&$ 0.00 \pm 0.00 $&$ 0.00 \pm 0.00  $&$ 0.00 \pm 0.00 $&$ 0.00 \pm 0.00  $&$ 0.00 \pm 0.00 $&$ 0.00 \pm 0.00  $&$ 0.00 \pm 0.00$ \\ \midrule
$\mathcal{T}(4,0,1)$     & T    &$ 21.3 \pm 22.7             $&$ 16.7 \pm 23.3^*            $&$ 9.11 \pm 16.0              $&$ 1.25 \pm 3.85^* $&$ 0.01 \pm 0.03 $&$ 0.00 \pm 0.01^* $&$ 0.00 \pm 0.00 $&$ 0.00 \pm 0.00^* $&$ 0.00 \pm 0.00 $&$ 0.00 \pm 0.00^* $&$ 0.00 \pm 0.00 $&$ 0.00 \pm 0.00^* $&$ 0.00 \pm 0.00 $&$ 0.00 \pm 0.00^* $& \textdagger   \\
$\cup$       & F                &$ 21.0 \pm 22.9             $&$ 16.8 \pm 23.6             $&$ 9.01 \pm 16.0              $&$ 1.26 \pm 3.74  $&$ 0.01 \pm 0.04 $&$ 0.00 \pm 0.01  $&$ 0.00 \pm 0.00 $&$ 0.00 \pm 0.00  $&$ 0.00 \pm 0.00 $&$ 0.00 \pm 0.00  $&$ 0.00 \pm 0.00 $&$ 0.00 \pm 0.00  $&$ 0.00 \pm 0.00 $&$ 0.00 \pm 0.00  $&$ 0.00 \pm 0.00$ \\
$\mathcal{T}(4,4,1)$    & B     &$ 21.0 \pm 22.9             $&$ 16.8 \pm 23.7             $&$ 8.87 \pm 15.9              $&$ 1.21 \pm 3.85  $&$ 0.00 \pm 0.03 $&$ 0.00 \pm 0.01  $&$ 0.00 \pm 0.01 $&$ 0.00 \pm 0.00  $&$ 0.00 \pm 0.00 $&$ 0.00 \pm 0.00  $&$ 0.00 \pm 0.00 $&$ 0.00 \pm 0.00  $&$ 0.00 \pm 0.00 $&$ 0.00 \pm 0.00  $&$ 0.00 \pm 0.00$ \\ \midrule
$\mathcal{L}(0,2)$       & T    &$ 24.1 \pm 24.9             $&$ 23.2 \pm 25.7^*            $&$ 16.4 \pm 23.3              $&$ 2.36 \pm 7.87^* $&$ 0.02 \pm 0.08 $&$ 0.00 \pm 0.00^* $&$ 0.00 \pm 0.00 $&$ 0.00 \pm 0.00^* $&$ 0.00 \pm 0.00 $&$ 0.00 \pm 0.00^* $&$ 0.00 \pm 0.00 $&$ 0.00 \pm 0.00^* $&$ 0.00 \pm 0.00 $&$ 0.00 \pm 0.00^* $& \textdagger   \\
$\cup$       & F                &$ 23.9 \pm 25.1             $&$ 23.3 \pm 26.0             $&$ 16.2 \pm 23.4              $&$ 2.37 \pm 7.82  $&$ 0.03 \pm 0.10 $&$ 0.00 \pm 0.01  $&$ 0.00 \pm 0.00 $&$ 0.00 \pm 0.00  $&$ 0.00 \pm 0.00 $&$ 0.00 \pm 0.00  $&$ 0.00 \pm 0.00 $&$ 0.00 \pm 0.00  $&$ 0.00 \pm 0.00 $&$ 0.00 \pm 0.00  $&$ 0.00 \pm 0.00$ \\
$\mathcal{L}(7,2)$       & B   &$ 23.8 \pm 24.8             $&$ 23.1 \pm 25.9             $&$ 16.4 \pm 23.5              $&$ 2.34 \pm 8.06  $&$ 0.02 \pm 0.10 $&$ 0.00 \pm 0.00  $&$ 0.00 \pm 0.01 $&$ 0.00 \pm 0.00  $&$ 0.00 \pm 0.00 $&$ 0.00 \pm 0.00  $&$ 0.00 \pm 0.00 $&$ 0.00 \pm 0.00  $&$ 0.00 \pm 0.00 $&$ 0.00 \pm 0.00  $&$ 0.00 \pm 0.00$ \\ \midrule
$\mathcal{U}(0,2)$       & T    &$ 0.07 \pm 0.09             $&$ 0.01 \pm 0.01^*            $&$ 0.00 \pm 0.00              $&$ 0.00 \pm 0.00^* $&$ 0.00 \pm 0.00 $&$ 0.00 \pm 0.00^* $&$ 0.00 \pm 0.00 $&$ 0.00 \pm 0.00^* $&$ 0.00 \pm 0.00 $&$ 0.00 \pm 0.00^* $&$ 0.00 \pm 0.00 $&$ 0.00 \pm 0.00^* $&$ 0.00 \pm 0.00 $&$ 0.00 \pm 0.00^* $& \textdagger   \\
$\cup$       & F               &$ 0.15 \pm 0.13             $&$ 0.04 \pm 0.03             $&$ 0.00 \pm 0.00              $&$ 0.00 \pm 0.00  $&$ 0.00 \pm 0.00 $&$ 0.00 \pm 0.00  $&$ 0.00 \pm 0.00 $&$ 0.00 \pm 0.00  $&$ 0.00 \pm 0.00 $&$ 0.00 \pm 0.00  $&$ 0.00 \pm 0.00 $&$ 0.00 \pm 0.00  $&$ 0.00 \pm 0.00 $&$ 0.00 \pm 0.00  $&$ 0.00 \pm 0.00$ \\
$\mathcal{U}(3,2)$        & B    &$ 0.08 \pm 0.13             $&$ 0.01 \pm 0.03             $&$ 0.00 \pm 0.01              $&$ 0.00 \pm 0.00  $&$ 0.00 \pm 0.00 $&$ 0.00 \pm 0.00  $&$ 0.00 \pm 0.00 $&$ 0.00 \pm 0.00  $&$ 0.00 \pm 0.00 $&$ 0.00 \pm 0.00  $&$ 0.00 \pm 0.00 $&$ 0.00 \pm 0.00  $&$ 0.00 \pm 0.00 $&$ 0.00 \pm 0.00  $&$ 0.00 \pm 0.00 $ \\ \midrule
$\mathcal{G}(2,-1,1)$    & T    &$ 1.00 \pm 5.80             $&$ 0.18 \pm 1.28^*            $&$ 0.01 \pm 0.04              $&$ 0.00 \pm 0.00^* $&$ 0.00 \pm 0.00 $&$ 0.00 \pm 0.00^* $&$ 0.00 \pm 0.00 $&$ 0.00 \pm 0.00^* $&$ 0.00 \pm 0.00 $&$ 0.00 \pm 0.00^* $&$ 0.00 \pm 0.00 $&$ 0.00 \pm 0.00^* $&$ 0.00 \pm 0.00 $&$ 0.00 \pm 0.00^* $& \textdagger   \\
$\cup$       & F                &$ 1.07 \pm 5.78             $&$ 0.21 \pm 1.23             $&$ 0.02 \pm 0.06              $&$ 0.00 \pm 0.00  $&$ 0.00 \pm 0.00 $&$ 0.00 \pm 0.00  $&$ 0.00 \pm 0.00 $&$ 0.00 \pm 0.00  $&$ 0.00 \pm 0.00 $&$ 0.00 \pm 0.00  $&$ 0.00 \pm 0.00 $&$ 0.00 \pm 0.00  $&$ 0.00 \pm 0.00 $&$ 0.00 \pm 0.00  $&$ 0.00 \pm 0.00$ \\
$\mathcal{G}(2,5,1)$     & B     &$ 0.98 \pm 5.48             $&$ 0.17 \pm 1.11             $&$ 0.01 \pm 0.06              $&$ 0.00 \pm 0.00  $&$ 0.00 \pm 0.00 $&$ 0.00 \pm 0.00  $&$ 0.00 \pm 0.00 $&$ 0.00 \pm 0.00  $&$ 0.00 \pm 0.00 $&$ 0.00 \pm 0.00  $&$ 0.00 \pm 0.00 $&$ 0.00 \pm 0.00  $&$ 0.00 \pm 0.00 $&$ 0.00 \pm 0.00  $&$ 0.00 \pm 0.00$ \\ \midrule
$\mathcal{E}(0,1)$       & T    &$ 0.08 \pm 0.36             $&$ 0.00 \pm 0.02^*            $&$ 0.00 \pm 0.00              $&$ 0.00 \pm 0.00^* $&$ 0.00 \pm 0.00 $&$ 0.00 \pm 0.00^* $&$ 0.00 \pm 0.00 $&$ 0.00 \pm 0.00^* $&$ 0.00 \pm 0.00 $&$ 0.00 \pm 0.00^* $&$ 0.00 \pm 0.00 $&$ 0.00 \pm 0.00^* $&$ 0.00 \pm 0.00 $&$ 0.00 \pm 0.00^* $& \textdagger   \\
$\cup$       & F                 &$ 0.11 \pm 0.37             $&$ 0.01 \pm 0.03             $&$ 0.00 \pm 0.01              $&$ 0.00 \pm 0.00  $&$ 0.00 \pm 0.00 $&$ 0.00 \pm 0.00  $&$ 0.00 \pm 0.00 $&$ 0.00 \pm 0.00  $&$ 0.00 \pm 0.00 $&$ 0.00 \pm 0.00  $&$ 0.00 \pm 0.00 $&$ 0.00 \pm 0.00  $&$ 0.00 \pm 0.00 $&$ 0.00 \pm 0.00  $&$ 0.00 \pm 0.00$ \\
$\mathcal{E}(4,1)$      & B      &$ 0.09 \pm 0.42             $&$ 0.00 \pm 0.01             $&$ 0.00 \pm 0.00              $&$ 0.00 \pm 0.00  $&$ 0.00 \pm 0.00 $&$ 0.00 \pm 0.00  $&$ 0.00 \pm 0.00 $&$ 0.00 \pm 0.00  $&$ 0.00 \pm 0.00 $&$ 0.00 \pm 0.00  $&$ 0.00 \pm 0.00 $&$ 0.00 \pm 0.00  $&$ 0.00 \pm 0.00 $&$ 0.00 \pm 0.00  $&$ 0.00 \pm 0.00$ \\ \midrule
$\mathcal{B}(2,2,1,1)$   & T     &$ 7.05 \pm 14.1             $&$ 4.69 \pm 10.4^*            $&$ 1.05 \pm 2.26              $&$ 0.03 \pm 0.13^* $&$ 0.00 \pm 0.00 $&$ 0.00 \pm 0.00^* $&$ 0.00 \pm 0.00 $&$ 0.00 \pm 0.00^* $&$ 0.00 \pm 0.00 $&$ 0.00 \pm 0.00^* $&$ 0.00 \pm 0.00 $&$ 0.00 \pm 0.00^* $&$ 0.00 \pm 0.00 $&$ 0.00 \pm 0.00^* $& \textdagger   \\
$\cup$       & F                 &$ 7.03 \pm 14.0             $&$ 4.72 \pm 10.4             $&$ 1.10 \pm 2.11              $&$ 0.05 \pm 0.16  $&$ 0.00 \pm 0.00 $&$ 0.00 \pm 0.00  $&$ 0.00 \pm 0.00 $&$ 0.00 \pm 0.00  $&$ 0.00 \pm 0.00 $&$ 0.00 \pm 0.00  $&$ 0.00 \pm 0.00 $&$ 0.00 \pm 0.00  $&$ 0.00 \pm 0.00 $&$ 0.00 \pm 0.00  $&$ 0.00 \pm 0.00$ \\
$\mathcal{B}(2,2,2,1)$   & B      &$ 6.80 \pm 14.0             $&$ 4.68 \pm 10.4             $&$ 1.00 \pm 2.07              $&$ 0.04 \pm 0.18  $&$ 0.00 \pm 0.00 $&$ 0.00 \pm 0.00  $&$ 0.00 \pm 0.00 $&$ 0.00 \pm 0.00  $&$ 0.00 \pm 0.00 $&$ 0.00 \pm 0.00  $&$ 0.00 \pm 0.00 $&$ 0.00 \pm 0.00  $&$ 0.00 \pm 0.00 $&$ 0.00 \pm 0.00  $&$ 0.00 \pm 0.00$ \\ \midrule
$\mathcal{T}_{nc}(4,2,0,1)$ & T    &$ 0.79 \pm 2.20             $&$ 0.32 \pm 2.17^*            $&$ 0.01 \pm 0.06              $&$ 0.00 \pm 0.00^* $&$ 0.00 \pm 0.00 $&$ 0.00 \pm 0.00^* $&$ 0.00 \pm 0.00 $&$ 0.00 \pm 0.00^* $&$ 0.00 \pm 0.00 $&$ 0.00 \pm 0.00^* $&$ 0.00 \pm 0.00 $&$ 0.00 \pm 0.00^* $&$ 0.00 \pm 0.00 $&$ 0.00 \pm 0.00^* $& \textdagger   \\
$\cup$       & F                   &$ 0.83 \pm 2.07             $&$ 0.37 \pm 2.16             $&$ 0.03 \pm 0.08              $&$ 0.00 \pm 0.00  $&$ 0.00 \pm 0.00 $&$ 0.00 \pm 0.00  $&$ 0.00 \pm 0.00 $&$ 0.00 \pm 0.00  $&$ 0.00 \pm 0.00 $&$ 0.00 \pm 0.00  $&$ 0.00 \pm 0.00 $&$ 0.00 \pm 0.00  $&$ 0.00 \pm 0.00 $&$ 0.00 \pm 0.00  $&$ 0.00 \pm 0.00$ \\
$\mathcal{T}_{nc}(4,2,7,1)$ & B     &$ 0.74 \pm 2.17             $&$ 0.36 \pm 2.42             $&$ 0.01 \pm 0.06              $&$ 0.00 \pm 0.00  $&$ 0.00 \pm 0.00 $&$ 0.00 \pm 0.00  $&$ 0.00 \pm 0.00 $&$ 0.00 \pm 0.00  $&$ 0.00 \pm 0.00 $&$ 0.00 \pm 0.00  $&$ 0.00 \pm 0.00 $&$ 0.00 \pm 0.00  $&$ 0.00 \pm 0.00 $&$ 0.00 \pm 0.00  $&$ 0.00 \pm 0.00$ \\ \midrule 
 \midrule
$\mathcal{N}(4,1)$       & T         &$ 19.0 \pm 23.1             $&$ 9.19 \pm 12.4^*            $&$ 7.03 \pm 13.7              $&$ 0.35 \pm 0.89^* $&$ 0.00 \pm 0.01 $&$ 0.00 \pm 0.00^* $&$ 0.00 \pm 0.00 $&$ 0.00 \pm 0.00^* $&$ 0.00 \pm 0.00 $&$ 0.00 \pm 0.00^* $&$ 0.00 \pm 0.00 $&$ 0.00 \pm 0.00^* $&$ 0.00 \pm 0.00 $&$ 0.00 \pm 0.00^* $& \textdagger   \\
$\cup$       & F                     &$ 18.8 \pm 23.2             $&$ 9.16 \pm 12.4             $&$ 6.92 \pm 13.6              $&$ 0.38 \pm 0.85  $&$ 0.01 \pm 0.02 $&$ 0.00 \pm 0.00  $&$ 0.00 \pm 0.00 $&$ 0.00 \pm 0.00  $&$ 0.00 \pm 0.00 $&$ 0.00 \pm 0.00  $&$ 0.00 \pm 0.00 $&$ 0.00 \pm 0.00  $&$ 0.00 \pm 0.00 $&$ 0.00 \pm 0.00  $&$ 0.00 \pm 0.00$ \\
$\mathcal{T}(4,0,1)$     & B        &$ 18.7 \pm 23.1             $&$ 9.35 \pm 12.6             $&$ 6.90 \pm 13.6              $&$ 0.33 \pm 0.88  $&$ 0.00 \pm 0.01 $&$ 0.00 \pm 0.00  $&$ 0.00 \pm 0.00 $&$ 0.00 \pm 0.00  $&$ 0.00 \pm 0.00 $&$ 0.00 \pm 0.00  $&$ 0.00 \pm 0.00 $&$ 0.00 \pm 0.00  $&$ 0.00 \pm 0.00 $&$ 0.00 \pm 0.00  $&$ 0.00 \pm 0.00$ \\ \midrule
$\mathcal{N}(4,1)$       & T         &$ 59.4 \pm 32.6             $&$ 55.5 \pm 32.0^*            $&$ 54.0 \pm 30.9              $&$ 44.7 \pm 32.9^* $&$ 25.4 \pm 28.0 $&$ 21.6 \pm 23.9^* $&$ 13.0 \pm 17.6 $&$ 0.89 \pm 3.11^* $&$ 0.01 \pm 0.05 $&$ 0.00 \pm 0.00^* $&$ 0.00 \pm 0.00 $&$ 0.00 \pm 0.00^* $&$ 0.00 \pm 0.00 $&$ 0.00 \pm 0.00^* $& \textdagger   \\
$\cup$       & F                          &$ 59.5 \pm 32.9             $&$ 56.0 \pm 32.3             $&$ 54.0 \pm 31.2              $&$ 44.7 \pm 33.2  $&$ 25.2 \pm 28.2 $&$ 21.4 \pm 24.1  $&$ 12.8 \pm 17.6 $&$ 0.94 \pm 3.05  $&$ 0.02 \pm 0.07 $&$ 0.00 \pm 0.01  $&$ 0.00 \pm 0.00 $&$ 0.00 \pm 0.00  $&$ 0.00 \pm 0.00 $&$ 0.00 \pm 0.00  $&$ 0.00 \pm 0.00$ \\
$\mathcal{L}(0,2)$       & B                          &$ 59.4 \pm 32.8             $&$ 56.0 \pm 32.3             $&$ 54.2 \pm 31.1              $&$ 44.6 \pm 33.2  $&$ 25.4 \pm 28.3 $&$ 21.4 \pm 23.9  $&$ 12.7 \pm 17.6 $&$ 0.85 \pm 2.91  $&$ 0.01 \pm 0.03 $&$ 0.00 \pm 0.00  $&$ 0.00 \pm 0.00 $&$ 0.00 \pm 0.00  $&$ 0.00 \pm 0.00 $&$ 0.00 \pm 0.00  $&$ 0.00 \pm 0.00$ \\ \midrule
$\mathcal{N}(4,1)$       & T                          &$ 24.2 \pm 26.4             $&$ 20.0 \pm 23.3^*            $&$ 11.2 \pm 18.4              $&$ 1.14 \pm 2.95^* $&$ 0.03 \pm 0.18 $&$ 0.00 \pm 0.04^* $&$ 0.00 \pm 0.00 $&$ 0.00 \pm 0.00^* $&$ 0.00 \pm 0.00 $&$ 0.00 \pm 0.00^* $&$ 0.00 \pm 0.00 $&$ 0.00 \pm 0.00^* $&$ 0.00 \pm 0.00 $&$ 0.00 \pm 0.00^* $& \textdagger   \\
$\cup$       & F                          &$ 24.0 \pm 26.6             $&$ 20.1 \pm 23.5             $&$ 12.1 \pm 18.4              $&$ 1.17 \pm 2.86  $&$ 0.04 \pm 0.19 $&$ 0.01 \pm 0.05  $&$ 0.00 \pm 0.00 $&$ 0.00 \pm 0.00  $&$ 0.00 \pm 0.00 $&$ 0.00 \pm 0.00  $&$ 0.00 \pm 0.00 $&$ 0.00 \pm 0.00  $&$ 0.00 \pm 0.00 $&$ 0.00 \pm 0.00  $&$ 0.00 \pm 0.00$ \\
$\mathcal{U}(0,2)$       & B                          &$ 23.8 \pm 26.5             $&$ 20.1 \pm 23.5             $&$ 11.0 \pm 18.4              $&$ 1.08 \pm 2.91  $&$ 0.03 \pm 0.16 $&$ 0.00 \pm 0.01  $&$ 0.00 \pm 0.00 $&$ 0.00 \pm 0.00  $&$ 0.00 \pm 0.00 $&$ 0.00 \pm 0.00  $&$ 0.00 \pm 0.00 $&$ 0.00 \pm 0.00  $&$ 0.00 \pm 0.00 $&$ 0.00 \pm 0.00  $&$ 0.00 \pm 0.00$ \\ \midrule
$\mathcal{N}(4,1)$       & T                          &$ 28.8 \pm 28.3             $&$ 28.2 \pm 28.3^*            $&$ 28.0 \pm 28.5              $&$ 9.39 \pm 16.1^* $&$ 3.37 \pm 8.94 $&$ 1.07 \pm 3.24^* $&$ 0.13 \pm 0.55 $&$ 0.00 \pm 0.00^* $&$ 0.00 \pm 0.00 $&$ 0.00 \pm 0.00^* $&$ 0.00 \pm 0.00 $&$ 0.00 \pm 0.00^* $&$ 0.00 \pm 0.00 $&$ 0.00 \pm 0.00^* $& \textdagger   \\
$\cup$       & F                          &$ 28.6 \pm 28.5             $&$ 28.4 \pm 28.7             $&$ 27.9 \pm 28.7              $&$ 9.33 \pm 16.1  $&$ 3.32 \pm 8.86 $&$ 1.05 \pm 3.05  $&$ 0.15 \pm 0.54 $&$ 0.00 \pm 0.00  $&$ 0.00 \pm 0.00 $&$ 0.00 \pm 0.00  $&$ 0.00 \pm 0.00 $&$ 0.00 \pm 0.00  $&$ 0.00 \pm 0.00 $&$ 0.00 \pm 0.00  $&$ 0.00 \pm 0.00$ \\
$\mathcal{G}(2,-1,1)$    & B                          &$ 28.6 \pm 28.5             $&$ 28.5 \pm 28.8             $&$ 28.0 \pm 28.6              $&$ 9.30 \pm 16.1  $&$ 3.31 \pm 8.94 $&$ 1.02 \pm 3.00  $&$ 0.13 \pm 0.56 $&$ 0.00 \pm 0.00  $&$ 0.00 \pm 0.00 $&$ 0.00 \pm 0.00  $&$ 0.00 \pm 0.00 $&$ 0.00 \pm 0.00  $&$ 0.00 \pm 0.00 $&$ 0.00 \pm 0.00  $&$ 0.00 \pm 0.00$ \\ \midrule
$\mathcal{N}(4,1)$       & T                          &$ 17.6 \pm 23.2             $&$ 9.91 \pm 14.5^*            $&$ 9.01 \pm 16.6              $&$ 1.26 \pm 4.10^* $&$ 0.05 \pm 0.33 $&$ 0.00 \pm 0.00^* $&$ 0.00 \pm 0.00 $&$ 0.00 \pm 0.00^* $&$ 0.00 \pm 0.00 $&$ 0.00 \pm 0.00^* $&$ 0.00 \pm 0.00 $&$ 0.00 \pm 0.00^* $&$ 0.00 \pm 0.00 $&$ 0.00 \pm 0.00^* $& \textdagger   \\
$\cup$       & F                          &$ 17.5 \pm 23.3             $&$ 9.94 \pm 14.6             $&$ 8.94 \pm 16.6              $&$ 1.29 \pm 4.03  $&$ 0.07 \pm 0.31 $&$ 0.00 \pm 0.01  $&$ 0.00 \pm 0.00 $&$ 0.00 \pm 0.00  $&$ 0.00 \pm 0.00 $&$ 0.00 \pm 0.00  $&$ 0.00 \pm 0.00 $&$ 0.00 \pm 0.00  $&$ 0.00 \pm 0.00 $&$ 0.00 \pm 0.00  $&$ 0.00 \pm 0.00$ \\
$\mathcal{E}(0,1)$       & B                          &$ 17.5 \pm 23.4             $&$ 9.91 \pm 14.7             $&$ 8.79 \pm 16.6              $&$ 1.26 \pm 4.00  $&$ 0.05 \pm 0.30 $&$ 0.00 \pm 0.01  $&$ 0.00 \pm 0.00 $&$ 0.00 \pm 0.00  $&$ 0.00 \pm 0.00 $&$ 0.00 \pm 0.00  $&$ 0.00 \pm 0.00 $&$ 0.00 \pm 0.00  $&$ 0.00 \pm 0.00 $&$ 0.00 \pm 0.00  $&$ 0.00 \pm 0.00$ \\ \midrule
$\mathcal{N}(4,1)$       & T                          &$ 22.3 \pm 25.4             $&$ 19.4 \pm 25.5^*            $&$ 10.3 \pm 15.5              $&$ 1.51 \pm 3.26^* $&$ 0.02 \pm 0.05 $&$ 0.00 \pm 0.00^* $&$ 0.00 \pm 0.00 $&$ 0.00 \pm 0.00^* $&$ 0.00 \pm 0.00 $&$ 0.00 \pm 0.00^* $&$ 0.00 \pm 0.00 $&$ 0.00 \pm 0.00^* $&$ 0.00 \pm 0.00 $&$ 0.00 \pm 0.00^* $& \textdagger   \\
$\cup$       & F                          &$ 22.1 \pm 25.5             $&$ 19.5 \pm 25.8             $&$ 10.1 \pm 15.4              $&$ 1.51 \pm 3.09  $&$ 0.03 \pm 0.07 $&$ 0.00 \pm 0.01  $&$ 0.00 \pm 0.00 $&$ 0.00 \pm 0.00  $&$ 0.00 \pm 0.00 $&$ 0.00 \pm 0.00  $&$ 0.00 \pm 0.00 $&$ 0.00 \pm 0.00  $&$ 0.00 \pm 0.00 $&$ 0.00 \pm 0.00  $&$ 0.00 \pm 0.00$ \\
$\mathcal{B}(2,2,1,1)$   & B                          &$ 22.1 \pm 25.4             $&$ 19.8 \pm 25.8             $&$ 10.3 \pm 15.4              $&$ 1.47 \pm 3.17  $&$ 0.01 \pm 0.05 $&$ 0.00 \pm 0.01  $&$ 0.00 \pm 0.00 $&$ 0.00 \pm 0.00  $&$ 0.00 \pm 0.00 $&$ 0.00 \pm 0.00  $&$ 0.00 \pm 0.00 $&$ 0.00 \pm 0.00  $&$ 0.00 \pm 0.00 $&$ 0.00 \pm 0.00  $&$ 0.00 \pm 0.00$ \\ \midrule
$\mathcal{N}(4,1)$       & T                          &$ 69.8 \pm 26.9             $&$ 67.7 \pm 27.0^*            $&$ 73.9 \pm 22.9              $&$ 82.8 \pm 19.0^* $&$ 82.9 \pm 20.5 $&$ 88.4 \pm 15.8^* $&$ 90.6 \pm 12.5 $&$ 95.5 \pm 7.56^* $&$ 96.9 \pm 8.89 $&$ 97.2 \pm 5.51^* $&$ 98.5 \pm 4.34 $&$ 99.1 \pm 2.97^* $&$ 99.7 \pm 0.44 $&$ 99.8 \pm 0.22^* $& \textdagger   \\
$\cup$       & F                          &$ 69.9 \pm 27.2             $&$ 68.3 \pm 27.2             $&$ 74.1 \pm 23.1              $&$ 83.1 \pm 19.0  $&$ 83.1 \pm 20.6 $&$ 88.7 \pm 15.8  $&$ 90.8 \pm 12.5 $&$ 95.7 \pm 7.54  $&$ 97.0 \pm 8.91 $&$ 97.4 \pm 5.52  $&$ 98.7 \pm 4.36 $&$ 99.3 \pm 2.97  $&$ 99.9 \pm 0.43 $&$ 1.00 \pm 0.13  $&$ 1.00 \pm 0.06$ \\
$\mathcal{T}_{nc}(4,2,0,1)$ & B                          &$ 69.8 \pm 27.1             $&$ 68.0 \pm 27.3             $&$ 73.7 \pm 23.4              $&$ 83.0 \pm 19.2  $&$ 83.2 \pm 20.3 $&$ 88.8 \pm 15.9  $&$ 90.8 \pm 12.6 $&$ 95.8 \pm 7.60  $&$ 97.0 \pm 9.11 $&$ 97.4 \pm 5.59  $&$ 98.8 \pm 4.29 $&$ 99.4 \pm 2.94  $&$ 99.9 \pm 0.36 $&$ 1.00 \pm 0.08  $&$ 1.00 \pm 0.02$ \\ \bottomrule
\end{tabular}}
\label{tab:multi_distr}
\end{table}

\end{landscape}